\documentclass[11pt]{article}

\usepackage[margin=1in]{geometry}
\usepackage{graphicx}
\usepackage{amsmath,amssymb}
\usepackage{booktabs}
\usepackage{multirow}
\usepackage{subcaption}
\usepackage{xcolor}
\usepackage{url}
\usepackage{xspace}
\usepackage{cite}
\usepackage[hidelinks]{hyperref}
\usepackage[capitalise,noabbrev]{cleveref}

\newcommand{\keywords}[1]{\par\small\noindent\textbf{Keywords:} #1\par}

\newcommand{\BGFG}{\ensuremath{\text{BG/FG}}}
\newcommand{\BGSFG}{\ensuremath{\text{BG+Sil/FG}}}
\newcommand{\mirsim}{\text{mirror\_sim}}
\newcommand{\nnsim}{\text{nn\_sim}}
\newcommand{\dangerm}{\text{danger\_margin}}
\newcommand{\eg}{e.g.\xspace}

\newcommand{\etal}{et~al.\xspace}
\DeclareMathOperator{\cosim}{cos}

\newcommand{\ArxivAuthors}{%
Antonio Rueda-Toicen$^{1,2}$, Abigail Allen Martin$^{3,4}$, Daniil Morozov$^{5,8}$, Matin Mahmood$^{8}$,\\
Alexandra Schild$^{1}$, Shahabeddin Dayani$^{6}$, Davide Panza$^{7}$, Gerard de Melo$^{1}$%
}
\newcommand{\ArxivAffiliations}{%
$^{1}$Hasso Plattner Institute \quad $^{2}$NVIDIA\\
$^{3}$Jaguar ID Project \quad $^{4}$Universidade Federal de Mato Grosso\\
$^{5}$Technical University Munich \quad $^{6}$Helmholtz Zentrum Berlin\\
$^{7}$Freie Universitaet Berlin \quad $^{8}$Hyper3Labs\\
\texttt{antonio.rueda-toicen@hpi.de}, \texttt{info@jaguaridproject.com}\\
\texttt{daniil.morozov@tum.de}, \texttt{aleksandra.kudaeva@hpi.de}\\
\texttt{gerard.demelo@hpi.de}%
}

\title{Are We Recognizing the Jaguar or Its Background?\\A Diagnostic Framework for Jaguar Re-Identification}
\author{\ArxivAuthors\\[0.5em]\normalsize \ArxivAffiliations}
\date{}

\begin{document}
\maketitle

% =====================================================================
% ABSTRACT
% =====================================================================
\begin{abstract}
Jaguar re-identification (re-ID) from citizen-science imagery can look strong
on standard retrieval metrics while still relying on the wrong evidence, such
as background context or silhouette shape, instead of the coat pattern that
defines identity. We introduce a diagnostic framework for wildlife re-ID with
two axes: a leakage-controlled context ratio,
\textit{background/foreground}, computed from inpainted background-only versus
foreground-only images, and a laterality diagnostic based on cross-flank
retrieval and mirror self-similarity. To make these diagnostics measurable, we
curate a Pantanal jaguar benchmark with per-pixel segmentation masks and an
identity-balanced evaluation protocol. We then use representative mitigation
families, ArcFace fine-tuning, anti-symmetry regularization, and Lorentz
hyperbolic embeddings, as case studies under the same evaluation lens. The
goal is not only to ask which model ranks best, but also what visual evidence
it uses to do so.
\end{abstract}

% =====================================================================
% 1  INTRODUCTION  (~1.5 pages)
% =====================================================================
\section{Introduction}\label{sec:intro}
% Target: ~1.5 pages

\begin{figure}[!t]
\centering
\includegraphics[width=0.95\textwidth]{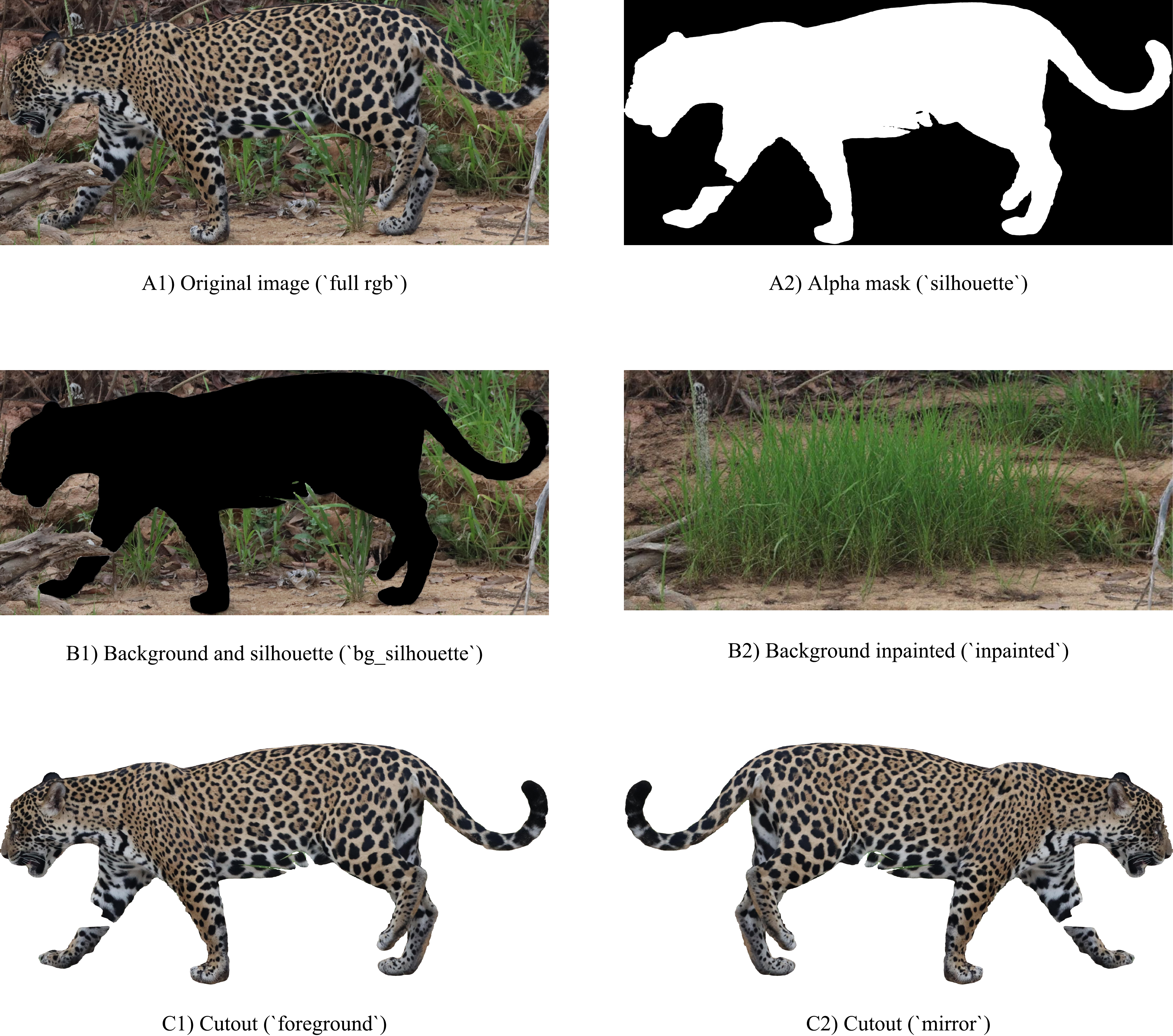}
\caption{\textbf{Six diagnostic image variants} derived from a single
citizen-science photograph using its SAM-3 alpha mask.
\textbf{(A1)}~Original \texttt{full\_rgb} input.
\textbf{(A2)}~Binary \texttt{silhouette} (alpha channel).
\textbf{(B1)}~\texttt{bg\_silhouette}: foreground replaced by a black
silhouette, retaining background context and shape cues.
\textbf{(B2)}~\texttt{inpainted}: foreground removed by FLUX.1-Fill
generative inpainting, eliminating the silhouette-shaped hole used for
leakage-controlled \BGFG{} evaluation (\cref{sec:bgfg}).
\textbf{(C1)}~\texttt{foreground} cutout (background removed), isolating
coat pattern.
\textbf{(C2)}~Horizontally flipped \texttt{mirror} cutout, used for
laterality analysis (\cref{sec:mirror}).
Jaguar rosette patterns are individually unique~\cite{jaguaridproject2025fieldguide};
the left flank differs from the right, so a mirrored image depicts an
individual that does not exist.}
\label{fig:fg_bg_decomposition}
\end{figure}

Citizen scientists and camera-trap networks now produce millions of wildlife images per year.
These data enable non-invasive population monitoring at large scale.
Individual re-identification (re-ID), the task of recognising specific
animals across sightings, is the foundation of mark-recapture studies,
movement ecology, and conservation management.  Deep learning has brought
high ranking accuracy to this task, yet accuracy alone can be
misleading: a model may achieve high scores by memorising the \emph{habitat}
rather than the \emph{animal}.  Geirhos~\etal~\cite{Geirhos2020shortcut}
formalised this pathology as \emph{shortcut learning}, and
Beery~\etal~\cite{Beery2018terra} demonstrated that camera-trap classifiers
exploit background context so much that they fail when deployed at new
sites.

Existing benchmarks for wildlife re-ID~\cite{vcermak2024wildlifedatasets,
cermak2024wildfusion, otarashvili2024multispecies} rank models by
accuracy-based metrics such as mean Average Precision (mAP) and Cumulative
Matching Characteristic (CMC), but few probe \emph{which visual features}
drive the match.  For species with unique bilateral coat patterns, jaguars
(\emph{Panthera onca}), leopards, whale sharks, an additional failure mode
exists. Recent sea-turtle work shows that opposite-side visual similarity can
sometimes be exploited to improve animal photo-identification, suggesting that
laterality effects are species-dependent rather than uniform across
taxa~\cite{adam2025exploiting}. The true coat pattern of these species is
\emph{asymmetric}: the rosette layout on the left flank differs from that on
the right~\cite{crall2013hotspotter}.  Models trained with horizontal
flip augmentation learn representations that collapse left and right flanks
into nearly identical embeddings.  More broadly, neural networks often learn
transformed copies of features under common image transformations, including
flips, making this kind of left-right equivalence a plausible
representational tendency rather than a jaguar-specific
artifact~\cite{olah2020naturally}. Yet a controlled experiment
(\cref{sec:crossflank}) shows that flips do not improve cross-flank
retrieval and reduce within-flank accuracy.  The relevant question is
whether a model can match the same individual across viewpoints; we call
this the \emph{laterality} axis and measure it via cross-flank retrieval
and mirror self-similarity.

These two failure modes, background shortcut and laterality shortcut, need not
move together.  Across our frozen baseline suite we do not see a clear monotonic
association between context reliance (\BGFG{}) and mean mirror similarity
(Axis~2; Sec.~\ref{sec:independence}), motivating a \emph{multi-axis}
diagnostic and mitigation framework.

This paper is about how to evaluate wildlife re-ID systems when shortcut cues
can dominate retrieval; the benchmark and mitigation families are included to
make that evaluation concrete.

Our contributions are as follows:

\begin{enumerate}
\item \textbf{Diagnostic Framework (C1).}  We propose two diagnostic axes for
  wildlife re-ID with segmentation masks: \BGFG{}, a leakage-controlled measure
  of context reliance, and Mirror Similarity, a laterality diagnostic for
  bilaterally asymmetric species defined on foreground-only cutouts with each
  model's native retrieval score.  We also analyse how the two axes relate in
  Sec.~\ref{sec:independence}.

\item \textbf{Jaguar Benchmark with Segmentation Masks (C2).}  We release a
  citizen-science Pantanal jaguar dataset with 1{,}895 RGBA images,
  per-pixel segmentation masks, and 31 identified individuals, and establish a
  reproducible evaluation protocol via two public Kaggle challenges:
  (\href{https://www.kaggle.com/competitions/jaguar-re-id}{Round 1})
  and
  (\href{https://www.kaggle.com/competitions/round-2-jaguar-reidentification-challenge}{Round 2}),
  with identity-balanced mAP as the official metric. Through these two public
  leaderboards, we explore the performance difference between models that use
  full RGB images and models that use foreground-only cutouts.
  The dataset is publicly available on Hugging Face
  (\href{https://huggingface.co/datasets/jaguaridentification/jaguars}{dataset})
  and is sourced from the Jaguar ID Project, a citizen-science initiative that
  collects and curates jaguar sightings across the Brazilian Pantanal; more
  information about the project and the dataset can be found on the Jaguar ID
  Project website
  (\href{https://www.jaguaridproject.com/}{Jaguar ID Project website}).

\item \textbf{Empirical Audit of Mitigation Families (C3).}  We use
  representative mitigation families, ArcFace fine-tuning, anti-symmetry
  regularization, and Lorentz hyperbolic embeddings, as case studies under the
  two-axis diagnostic framework, and report their accuracy--robustness
  trade-offs.
\end{enumerate}

% =====================================================================
% 2  RELATED WORK  (~1.5 pages)
% =====================================================================
\section{Related Work}\label{sec:related}
% Target: ~1.5 pages

\subsection{Animal Re-Identification}

Classical wildlife re-ID relies on local keypoint descriptors.
HotSpotter~\cite{crall2013hotspotter} matches SIFT keypoints on patterned
species (\eg zebras, jaguars) by verifying spatial correspondences.
Deep global descriptors have become the dominant approach.
MiewID~\cite{otarashvili2024multispecies} trains a multi-species re-ID model
on a large community-curated dataset, producing embeddings tailored to
wildlife.  The WildlifeDatasets toolkit~\cite{vcermak2024wildlifedatasets}
standardises data loading and evaluation across dozens of species, while
WildFusion~\cite{cermak2024wildfusion} fuses multiple similarity scores
with calibrated confidence weighting.  Recent work on sea turtles studies the
cross-side problem directly, showing that left and right facial profiles retain
exploitable identity signal and that searching across both sides can improve
retrieval in that setting~\cite{adam2025exploiting}. Foundation models pre-trained on
large-scale data, including DINOv3~\cite{simeoni2025dinov3},
EVA-02~\cite{fang2024eva}, EfficientNetV2~\cite{tan2021efficientnetv2},
I-JEPA~\cite{assran2023ijepa},
ConvNeXtV2~\cite{woo2023convnextv2},
C-RADIO~\cite{ranzinger2026cradio},
and ResNet~\cite{he2016deep}, are now used as frozen or fine-tuned
backbones.  None of these works, however, probe
\emph{which image regions} drive the re-ID decision.

\subsection{Shortcut Learning and Spurious Correlations}

Geirhos~\etal~\cite{Geirhos2020shortcut} provide a taxonomy of shortcut
learning in deep networks, showing that models exploit texture,
context, or other spurious cues instead of the intended discriminative
features.  In the camera-trap domain,
Beery~\etal~\cite{Beery2018terra} showed that species classifiers memorise
background scenes so well that accuracy drops at unseen
sites, a finding we extend to the \emph{re-identification} setting.
Sagawa~\etal~\cite{Sagawa2020dro} propose distribution-robust
optimisation to improve worst-group performance under spurious correlations,
a principle that motivates our identity-balanced evaluation protocol.

\subsection{Metric Learning for Fine-Grained Recognition}

ArcFace~\cite{deng2019arcface} and its Sub-Center variant~\cite{deng2020sub} are angular margin
losses that produce discriminative embeddings while tolerating noisy
labels via multiple sub-centers per class.  Hyperbolic
embeddings~\cite{khrulkov2020hyperbolic, becigneul2019riemannian} offer an
alternative geometry that we use to focus on individually distinctive jaguar
flank patterns~\cite{jaguaridproject2025fieldguide} in foreground cutouts
rather than background or silhouette cues, with a mask-solidity hierarchy
that keeps uncertain, fragmented observations near the origin and pushes
high-solidity coat-pattern views outward as more specific embeddings
(\cref{sec:supp_lorentz_loss}).

\subsection{Local Feature Matching and Preprocessing}

Local feature methods still matter for patterned species.  Classical SIFT or
RootSIFT descriptors, aggregated with VLAD~\cite{jegou2010vlad}, provide
strong non-learned baselines.  Learned alternatives include
ALIKED~\cite{zhao2023aliked}, a lightweight deformable keypoint method, and
LoFTR~\cite{sun2021loftr}, a detector-free transformer matcher.  For
segmentation, SAM~3~\cite{carion2025sam} can produce per-pixel masks at scale.
Those masks enable the foreground/background decomposition central to our
Axis~1 diagnostic.  Histogram equalisation
techniques~\cite{hummel1975image}, such as Contrast Limited Adaptive
Histogram Equalisation (CLAHE), remain standard preprocessing for normalising
camera-trap lighting variations.

% =====================================================================
% 3  METHOD  (~3.5 pages)
% =====================================================================
\section{Diagnostic Framework and Benchmark}\label{sec:method}
% Target: ~3.5 pages

% ------------------------------------------------------------------
\subsection{Dataset and Benchmark}\label{sec:dataset}

We curate and release a jaguar re-ID dataset sourced from
citizen-science surveys in the Brazilian Pantanal.  The dataset comprises
\textbf{1{,}895 training images} across \textbf{31 individual jaguars} (13
to 183 images per individual, mean 61.1, imbalance ratio $14.1\times$) and
\textbf{371 test images} forming 137{,}270 query--gallery pairs.  All images
are provided in RGBA PNG format, where the alpha channel encodes a per-pixel
segmentation mask that separates the jaguar (foreground) from the background.
We generate these masks with SAM~3~\cite{carion2025sam} using the text prompt
\texttt{"jaguar"} and store the resulting binary mask as the alpha channel in
the released RGBA PNGs.  We do not manually edit masks; instead we quantify
mask quality with the \emph{mask solidity} $s \in [0,1]$, defined as the ratio
of mask area to convex-hull area (see Appendix \cref{sec:supp_masks}).
Image sizes vary from $193{\times}176$ to $6{,}932{\times}4{,}616$ pixels
(mean ${\sim}3{,}199{\times}2{,}065$), due to diverse camera setups.

\paragraph{Split construction and near-duplicate filtering.}
The held-out test split was constructed via stratified sampling to ensure
each identity is represented by at least $n{=}3$ images and to maintain
coverage of three coarse viewpoints (left flank, right flank, and frontal
views) when available.  To reduce near-duplicate leakage, we deduplicated
the dataset prior to splitting using frozen MegaDescriptor-L
embeddings~\cite{vcermak2024wildlifedatasets} computed on foreground-only
cutouts: image pairs with cosine similarity $> 0.8$ were treated as
near-duplicates and one instance was discarded.

These segmentation masks enable our diagnostic evaluation:
they let us deterministically generate six image variants per photograph
(\cref{fig:fg_bg_decomposition}), each isolating a different visual cue
(coat pattern, silhouette shape, background context, or laterality).

To standardise evaluation, we host two public Kaggle challenges, Round 1
(\href{https://www.kaggle.com/competitions/jaguar-re-id}{Round 1 Kaggle challenge})
and Round 2
(\href{https://www.kaggle.com/competitions/round-2-jaguar-reidentification-challenge}{Round 2 Kaggle challenge}),
with \textbf{identity-balanced mAP} (macro averaging) as the official
metric.  This prevents identities with many images from dominating the
score, critical given the $14.1\times$ class imbalance.  For $I$
identities:
\begin{equation}\label{eq:macro_map}
  \text{mAP}_{\text{macro}}
  = \frac{1}{I}\sum_{i=1}^{I}\,
    \frac{1}{|Q_i|}\sum_{q \in Q_i} \text{AP}(q)\,,
\end{equation}
where $Q_i$ is the set of queries belonging to identity~$i$.  We report both
macro (identity-balanced) and micro (instance-weighted) variants throughout.

\paragraph{Query--gallery protocol and CMC.}
The benchmark provides an explicit list of query--gallery pairs: each of the
371 test images is used as a query and is compared against all other test
images in the gallery (370 per query), yielding 137{,}270 ordered pairs with
self-matches excluded.  For each query $q$, gallery images are ranked by
predicted similarity (higher is better), and $\text{AP}(q)$ uses the binary
relevance label $\text{id}(q)=\text{id}(g)$ for a gallery image $g$.  We also report the Cumulative
Matching Characteristic (CMC) at rank $K$:
\begin{equation}\label{eq:cmc_micro}
  \text{CMC}_{\text{micro}}@K
  = \frac{1}{|Q|}\sum_{q \in Q} [\,\text{rank}(q) \le K\,]\,,
\end{equation}
where $Q$ is the set of all queries, $\text{rank}(q)$ is the (1-indexed) rank
of the first correct match, and $[\,\cdot\,]$ is 1 if the predicate is true and
0 otherwise.  Macro CMC averages per-identity CMC@K values, analogous to
\cref{eq:macro_map}.  For Lorentz methods we rank by hyperbolic distance and
map distances to Kaggle-format submission similarities via $\exp(-d)$; this
monotonic transform preserves the retrieval ordering.

% ------------------------------------------------------------------
\subsection{Diagnostic Axis~1: Background Context Ratio (\BGFG{})}\label{sec:bgfg}

Foreground/background decomposition enables controlled shortcut stress tests,
but the construction matters.  A na\"ive \texttt{background\_only} variant
that zeros foreground pixels leaves a jaguar-shaped silhouette hole and
boundary artifacts, which can encode non-coat information (pose, cropping,
camera framing) and confound a purely ``background-only'' interpretation.

\paragraph{Leakage-controlled background ratio via inpainting (\BGFG{}).}
To measure background reliance without silhouette leakage, we compute a
\texttt{background\_inpainted} variant where the removed foreground region is
filled with plausible background content using FLUX.1-Fill-dev
(details in \cref{sec:supp_inpainting}).  The diagnostic ratio is:
\begin{equation}\label{eq:bgfg_inpainted}
  \BGFG = \frac{\text{mAP}(\texttt{background\_inpainted})}
               {\text{mAP}(\texttt{foreground\_only})}\,,
\end{equation}
where both mAP values use the identity-balanced (macro) formulation of
\cref{eq:macro_map}.  Lower \BGFG{} indicates lower reliance on non-coat
context.  We additionally report \BGSFG{}, computed with the zeroed-foreground
\texttt{background\_only} construction for comparison; because that input
retains a jaguar-shaped hole and boundary cues, it measures background plus
silhouette leakage rather than pure background alone
(\cref{tab:supp_bgfg_inpainted}).

% ------------------------------------------------------------------
\subsection{Diagnostic Axis~2: Mirror Similarity}\label{sec:mirror}

Jaguar coat patterns are bilaterally asymmetric: the rosette layout on the
left flank differs from the right.  A horizontal flip of a left-flank image
therefore is not a real observation of that animal's right flank.  The animal
identity is unchanged, but the flip removes biologically meaningful laterality
and can encourage false left-right equivalence in the embedding space.  Our
biological endpoint for laterality robustness is cross-flank retrieval,
defined in Sec.~\ref{sec:crossflank} as matching the same individual across
opposite flanks.  Throughout the paper, however, Axis~2 refers to mean mirror
similarity on foreground-only cutouts under each model's native retrieval
score.  We use it as a scalable proxy for laterality collapse, while
nearest-neighbour similarity and danger margin serve as auxiliary per-image
diagnostics.  The mirrored-query retrieval ratio is reported only as a
secondary stress test.

Let $f(\cdot)$ denote a model's embedding function and $\text{flip}(x)$ the
horizontal mirror of image~$x$.  Let $s_\theta(a,b)$ denote the model's
retrieval score between images $a$ and $b$, with larger values meaning greater
similarity.  For Euclidean models we use
$s_\theta(a,b)=\cosim\!\bigl(f(a), f(b)\bigr)$ on $\ell_2$-normalised
embeddings.  For Lorentz models we use
$s_\theta(a,b)=\exp\!\bigl(-d_L(f(a), f(b))\bigr)$, where $d_L$ is Lorentz
geodesic distance; this matches the monotonic transform used for test-set
ranking in the query--gallery protocol above.  We define:
\begin{align}
  \mirsim(x)    &= s_\theta\bigl(x,\; \text{flip}(x)\bigr)\,,
    \label{eq:mirror_sim} \\
  \nnsim(x)     &= \max_{y:\,\text{id}(y)\neq\text{id}(x)}
                    s_\theta\bigl(\text{flip}(x),\; y\bigr)\,,
    \label{eq:nn_sim} \\
  \dangerm(x)   &= \nnsim(x) - \mirsim(x)\,.
    \label{eq:danger}
\end{align}

When $\dangerm(x) > 0$, the flipped image receives a higher retrieval score
against a \emph{different individual} than against the original
(\cref{fig:supp_asymmetry_bias} shows two such cases).  We classify
models into four tiers based on their mean mirror similarity across the test
set:

\begin{itemize}
\item \textbf{Tier~1: Laterality-Aware} ($< 0.85$): MiewID variants~\cite{otarashvili2024multispecies}.
\item \textbf{Tier~2: Moderate Shortcut} ($0.85$--$0.96$): MegaDescriptor~\cite{vcermak2024wildlifedatasets}, ResNet50~\cite{he2016deep}, DINOv3-ViT CLS~\cite{simeoni2025dinov3}.
\item \textbf{Tier~3: Strong Shortcut} ($0.96$--$0.99$): I-JEPA~\cite{assran2023ijepa}, DINOv3-Pooled, EfficientNetV2~\cite{tan2021efficientnetv2}, ConvNeXtV2~\cite{woo2023convnextv2}.
\item \textbf{Tier~4: Near-Perfect Symmetry} ($> 0.99$): C-RADIO~\cite{ranzinger2026cradio}, EVA-02~\cite{fang2024eva}, HotSpotter+Mean~\cite{crall2013hotspotter}.
\end{itemize}

\subsubsection{Relationship Between the Two Axes.}\label{sec:independence}

We test whether context reliance predicts laterality collapse by correlating
Axis~1 (\BGFG{}) with Axis~2, defined throughout as mean mirror similarity on
foreground-only cutouts under each model's native retrieval score.  Lower mean
mirror similarity indicates stronger laterality awareness because an image and
its horizontal flip receive a lower self-match score.  We keep the
mirrored-query retrieval ratio $r = \text{mAP}_{\text{mirror}} /
\text{mAP}_{\text{regular}}$ only as an auxiliary retrieval stress test in the
appendix.

Across the $N{=}15$ deep frozen models in Table~\ref{tab:frozen} (excluding the
classical HotSpotter+VLAD baseline),
the rank correlation between \BGFG{} and mean mirror similarity is weak:
\begin{equation}
  \text{Spearman}\;\rho = 0.307\;(p = 0.265)\,,\qquad
  \text{Kendall}\;\tau  = 0.181\;(p = 0.379)\,.
\end{equation}
With $N{=}15$, uncertainty is wide (bootstrap 95\% CI for $\rho$:
$[-0.360,\,0.771]$ and for $\tau$: $[-0.327,\,0.600]$; percentile CI with
$B{=}20{,}000$, seed~0), so we report this as \emph{no clear monotonic
association}, not evidence of independence.

Concrete counter-examples show why both axes must be measured:
\begin{itemize}
\item \textbf{EVA-02}: low \BGFG{} ($0.66$) but near-perfect mirror symmetry (mean mirror similarity $0.997$).
\item \textbf{I-JEPA}: high \BGFG{} ($1.17$) yet lower mean mirror similarity ($0.968$) than most self-supervised baselines.
\item \textbf{MiewID-MSv2}: only moderate \BGFG{} ($0.52$) but by far the strongest laterality awareness (mean mirror similarity $0.752$).
\end{itemize}
A complete robustness assessment therefore requires measuring \emph{both}
axes; excelling on one provides no guarantee about the other.

% ------------------------------------------------------------------
\subsection{Mitigation Families Used as Case Studies}\label{sec:mitigation}

We treat the mitigation methods in this paper as case studies under a common
diagnostic lens, not as co-equal method contributions.  Full objectives,
training details, and benchmark sweeps are collected in the appendix.

\subsubsection{ArcFace Fine-Tuning.}\label{sec:arcface}

ArcFace is our Euclidean fine-tuning baseline: we freeze a pre-trained
backbone, train a Sub-Center ArcFace head on foreground-only cutouts, and
omit horizontal flip augmentation.  A controlled ablation
(\cref{sec:crossflank}) shows that flips do not improve cross-flank
re-ID and reduce within-flank accuracy; we therefore omit them and evaluate
cross-flank performance explicitly.
The loss definition and training details are given in
\cref{sec:supp_euclidean_obj,sec:supp_arcface_bench}.

\subsubsection{Anti-Symmetry regularization.}\label{sec:antisym}

Anti-symmetry regularization targets Axis~2 directly by treating an image and
its horizontal flip as a soft negative pair.  We combine it with ArcFace to
test whether laterality-aware training improves retrieval without reintroducing
background shortcuts; the exact objective and ablation grid are reported in
\cref{sec:supp_euclidean_obj,sec:supp_antisym_ablation}.

\subsubsection{Three-Tier Negative Mining.}\label{sec:mining}

Our Euclidean training setup also uses three negative types motivated by the
diagnostics: mirrored cutouts, background cutouts, and hard different-individual
pairs from offline ranking.  This keeps the training signal aligned with the
same shortcut structure that we measure at evaluation time; the full mining
table is in \cref{sec:supp_mining}.

\subsubsection{Lorentz Hyperbolic Embeddings.}\label{sec:lorentz}

As a complementary case-study family, we evaluate Lorentz embeddings with a
solidity-conditioned radius prior and an optional mirror-negative term.  This
tests whether a different embedding geometry changes the
accuracy--robustness trade-off under the same framework; the objective,
training details, and validation benchmark are in
\cref{sec:supp_lorentz,sec:supp_lorentz_loss}.

% =====================================================================
% 4  EXPERIMENTS  (~5.0 pages)
% =====================================================================
\section{Empirical Audit}\label{sec:experiments}
% Target: ~5.0 pages

% ------------------------------------------------------------------
\subsection{Frozen Baseline Benchmark}\label{sec:frozen}

We evaluate 16 baselines without fine-tuning on three inputs: the original full
RGB images (baseline accuracy), foreground-only cutouts, and leakage-controlled
inpainted backgrounds.  This yields 48 test-set evaluations (16 models $\times$
3 variants).  \Cref{tab:frozen} reports identity-balanced (macro) mAP on the full
RGB images and both diagnostic axes: \BGFG{}, computed from the
inpainted-background and foreground-only mAPs (Eq.~\ref{eq:bgfg_inpainted}), and
mean mirror similarity computed on foreground-only images (Eq.~\ref{eq:mirror_sim}).
For completeness, the zeroed-foreground \BGSFG{} ratio, which retains
silhouette-hole cues, is reported in \cref{tab:supp_bgfg_inpainted}.
For the frozen deep backbones, embeddings are extracted with each backbone's
canonical preprocessing and a single global descriptor (CNN avgpool or ViT
token/pooled features), $\ell_2$-normalised, and scored with cosine
similarity, which is the native retrieval score used for Axis~2 in this
section.  Complete extraction details are provided in
\cref{sec:supp_compute,tab:supp_frozen_extraction}.

\begin{table}[t]
\centering
\caption{Frozen baseline benchmark.  16 models evaluated on three variants
(\texttt{full\_rgb}, \texttt{foreground\_only}, \texttt{background\_inpainted}),
yielding 48 test-set evaluations.  The mAP column is identity-balanced (macro)
test mAP on \texttt{full\_rgb}.  \BGFG{} uses
\texttt{background\_inpainted}/\texttt{foreground\_only}
(Eq.~\ref{eq:bgfg_inpainted}); mirror similarity is mean \mirsim on
\texttt{foreground\_only} (Eq.~\ref{eq:mirror_sim}).  Models are ranked by mAP.}
\label{tab:frozen}
\resizebox{\textwidth}{!}{%
\begin{tabular}{@{}rlcccccl@{}}
\toprule
Rank & Model & mAP (macro) & \BGFG{} & Mirror Sim & Mirror Tier & Architecture \\
\midrule
1  & MiewID-MSv3             & 0.303 & 0.59 & 0.746 & T1: Aware     & EffNetV2-M \\
2  & DINOv3-ViT CLS          & 0.298 & 0.77 & 0.963 & T2: Moderate  & ViT-S/16 \\
3  & MiewID-MSv2             & 0.297 & 0.52 & 0.752 & T1: Aware     & EffNetV2-M \\
4  & HotSpotter+VLAD         & 0.288 & 0.54 & 1.000 & T4: Symmetric & SIFT+VLAD \\
5  & EVA-02                  & 0.281 & 0.66 & 0.997 & T4: Symmetric & ViT-L/14 \\
6  & DINOv3-ViT Pooled       & 0.277 & 0.82 & 0.979 & T3: Strong    & ViT-S/16 \\
7  & C-RADIO-v4              & 0.255 & 0.78 & 0.997 & T4: Symmetric & ViT-CPE \\
8  & MegaDescriptor          & 0.249 & 0.77 & 0.914 & T2: Moderate  & ViT \\
9  & I-JEPA                  & 0.238 & 1.17 & 0.968 & T3: Strong    & ViT-H/14 \\
10 & ResNet50                & 0.218 & 0.90 & 0.958 & T2: Moderate  & CNN \\
11 & ResNet152               & 0.218 & 0.94 & 0.961 & T3: Strong    & CNN \\
12 & ConvNeXtV2-Base         & 0.210 & 1.00 & 0.985 & T3: Strong    & CNN \\
13 & DINOv3-ConvNeXt Pooled  & 0.209 & 0.92 & 0.981 & T3: Strong    & ConvNeXt-S \\
14 & DINOv3-ConvNeXt CLS     & 0.208 & 0.91 & 0.984 & T3: Strong    & ConvNeXt-S \\
15 & EfficientNetV2-M        & 0.200 & 1.03 & 0.981 & T3: Strong    & CNN \\
16 & meFeM-B                 & 0.186 & 1.12 & 0.970 & T3: Strong    & ViT-B \\
\bottomrule
\end{tabular}}
\end{table}

\paragraph{Frozen-model takeaways.}
Under the leakage-controlled \BGFG{} diagnostic, 3 of 14 frozen models have
$\BGFG \geq 1.0$ (EfficientNetV2-M, meFeM-B, I-JEPA), meaning they extract
more identity signal from background context than from coat pattern alone.
MiewID-MSv2, pre-trained on wildlife data~\cite{otarashvili2024multispecies},
achieves the lowest ratio at 0.52.  High frozen mAP does not guarantee low
shortcut reliance: DINOv3-ViT CLS reaches 0.298 macro mAP with \BGFG{} 0.77.
Across the broader 102-model suite, only three models exceed $\BGFG \ge 1.0$
and only two exceed 1.10, so the strongest context shortcuts are concentrated
in a small subset of encoders.

\paragraph{Architecture effects.}\leavevmode\\
Wildlife-specific pre-training matters.  MiewID-MSv2 and the generic
EfficientNetV2-M share the same backbone.  Their \BGFG{} values still differ
sharply (0.52 vs.\ 1.03).  Pooling also amplifies shortcuts for
DINOv3-ViT-S/16, raising \BGFG{} from 0.77 to 0.82 and mirror similarity
from 0.963 to 0.979.
Self-supervised models remain largely laterality-agnostic (mirror similarity
$> 0.96$), and the full HotSpotter comparison appears in
\cref{sec:supp_hotspotter,tab:supp_hotspotter}.

\begin{figure}[t]
\centering
\includegraphics[width=\textwidth]{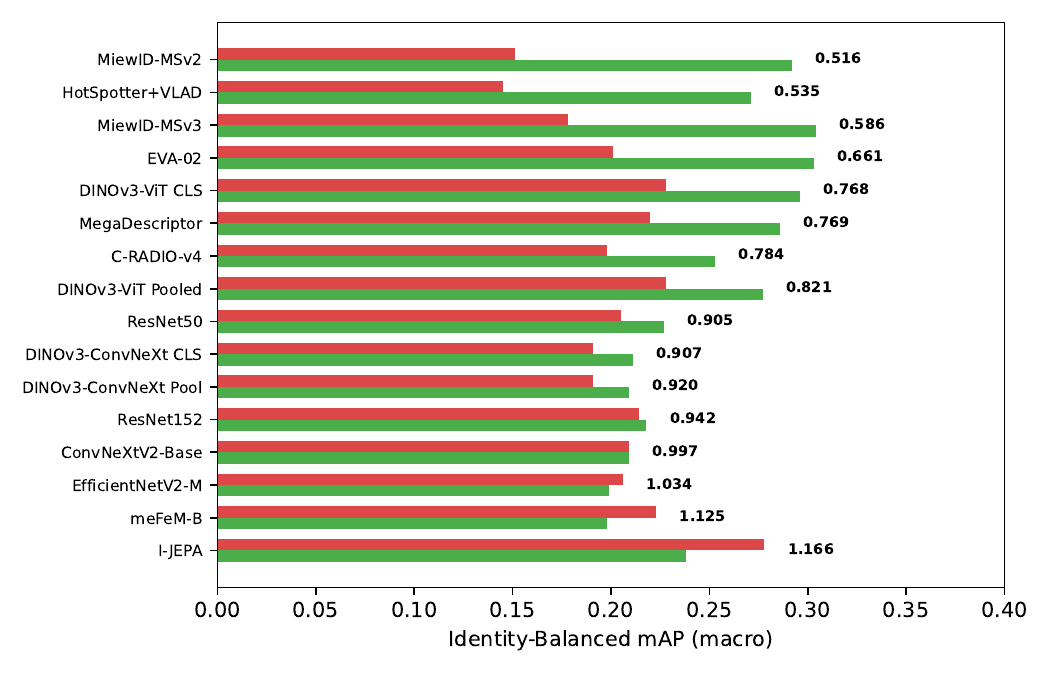}
\caption{\textbf{Foreground vs.\ inpainted background mAP across frozen models.}
Green bars show foreground-only mAP; red bars show inpainted background-only mAP.
Models are sorted by \BGFG{}, shown in \textbf{bold} next to each bar pair.
MiewID-MSv2 achieves the lowest frozen \BGFG{} (0.52), consistent with
wildlife-specific pre-training reducing non-coat context reliance.}
\label{fig:bgfg_bar}
\end{figure}

\subsection{Euclidean Case Studies}\label{sec:arcface_results}

We use ArcFace fine-tuning as the main Euclidean case-study family.  On
foreground-only training, the active matched ArcFace pair lands close on the
test benchmark.  O0 is slightly higher on full-RGB macro mAP (0.483
vs.\ 0.469), while O1 is slightly higher on cross-flank CMC@1 (62.4\% vs.\
60.7\%).  Those gaps are descriptive, not decisive.  After the paired
Wilcoxon-Fisher-Holm pipeline, the retrieval endpoints between ArcFace O0 and
O1 are not supported.  The one supported ArcFace difference is on Axis~2
danger margin, where O0 is lower.  That is why we keep the diagnostic
axes separate from the leaderboard: the benchmark tells us where the models
land, while the paired tests tell us which pairwise differences are sturdy
enough to claim.

% ------------------------------------------------------------------
\subsection{Shortcut Axis Analysis}\label{sec:mirror_results}

We rank all 15 models by mean mirror similarity on foreground-only test
cutouts.  This frozen ranking contains no Lorentz models, so Axis~2 reduces
to each descriptor model's native similarity score; for the deep baselines
that score is cosine on the extracted embeddings.  Wildlife pre-training is
the primary determinant of laterality awareness: the gap between MiewID
($0.752$) and the next-best non-MiewID model (MegaDescriptor, $0.914$) is
$0.162$, larger than the entire spread among non-MiewID models.  ArcFace
fine-tuning degrades laterality (MiewID frozen $0.752 \to$ ArcFace+MiewID
$0.833$), which motivated the anti-symmetry regularization of
\cref{sec:antisym}.  The full 15-model ranking and laterality sensitivity
visualisation are in \cref{tab:supp_mirror,fig:supp_laterality_ratio}.

\paragraph{MegaDescriptor training-set scan.}
On the 1{,}895 training images, a full danger-margin scan with MegaDescriptor
finds only 2~images (0.1\%) with $\dangerm > 0$, both from low-contrast
identities (Overa, Medrosa).  The mean danger margin is $-0.366$ (median
$-0.369$).  Per-identity statistics are in \cref{tab:supp_danger_margin}.

\begin{figure}[t]
\centering
\includegraphics[width=\textwidth]{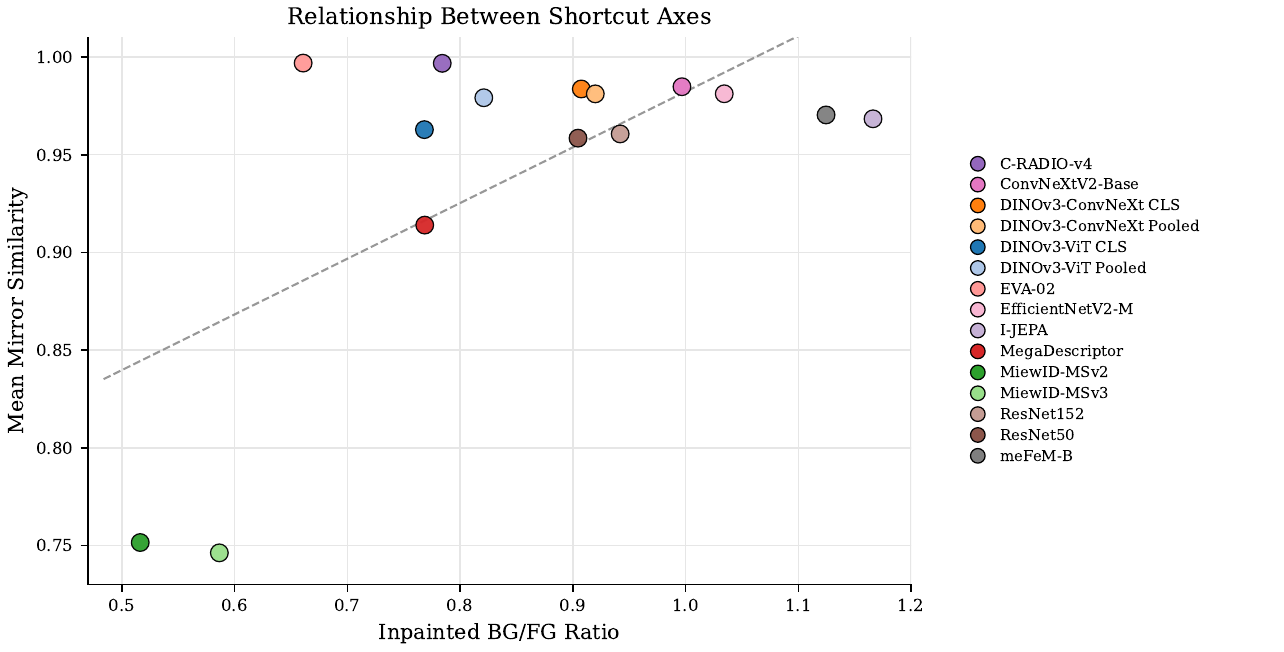}
\caption{\textbf{Relationship between shortcut axes.}  Each point is a frozen
model plotted by inpainted \BGFG{} ($x$) and mean mirror similarity ($y$),
both computed on foreground-only cutouts (lower mean mirror similarity
indicates greater laterality awareness).  Spearman $\rho = 0.307$
($p = 0.265$; $N{=}15$; 95\% bootstrap CI $[-0.360,\,0.771]$,
$B{=}20{,}000$, seed~0): no clear monotonic association.  Counter-examples
include EVA-02 (\BGFG{} $0.661$, mean mirror similarity $0.997$) and I-JEPA
(\BGFG{} $1.166$, mean mirror similarity $0.968$).}
\label{fig:bgfg_mirror_scatter}
\end{figure}

\paragraph{Axis relationship.}
\Cref{fig:bgfg_mirror_scatter} plots inpainted \BGFG{} against the
mean mirror similarity across the frozen suite.
We do not see a clear monotonic relationship between the axes
(Sec.~\ref{sec:independence}).  Several models are robust on one axis but
brittle on the other, so we treat them as complementary diagnostics rather
than interchangeable summaries.

% ------------------------------------------------------------------
\subsection{Paired Objective Comparisons}\label{sec:ablations}

 The paired AP tests support four differences.  ArcFace O0 vs.\ O1 is supported on Axis~2
danger margin only.  Lorentz O0 vs.\ O1 (MSv3) has no supported AP or Axis~2
difference.  Triplet T0 vs.\ T1 is supported on the two inpainted vegetation AP
endpoints, plus Axis~2 danger margin and raw mirror similarity.  The paired CMC
tests are even more conservative: none of the matched pairs shows a supported
CMC@1/5/10 difference after Fisher combination and Holm correction.  The full
paired-test summaries appear in \cref{tab:supp_paired_ap,tab:supp_paired_cmc}.

% ------------------------------------------------------------------
\subsection{Cross-Flank Retrieval and Flip Augmentation Ablation}\label{sec:crossflank}

The biological target of laterality robustness is cross-flank matching:
retrieving the same individual when query and gallery show opposite flanks. 
We parse flank visibility labels from the 371 test descriptions, yielding
123 clean left-flank and 98 clean right-flank images across 23 identities
with both flanks represented.  We define two retrieval protocols on
foreground-only cutouts:
\emph{within-flank} (L$\to$L and R$\to$R, with self-exclusion) and
\emph{cross-flank} (L$\to$R and R$\to$L, disjoint query--gallery sets).
Queries without a positive in the relevant gallery are excluded from that
split.

To test whether flip augmentation helps cross-flank matching, we train a
second ArcFace+MiewID-MSv2 model identical to the no-flip baseline except
for adding \texttt{HorizontalFlip(p=0.5)} during training.
\Cref{tab:crossflank} reports the results.

\begin{table}[t]
\centering
\caption{Cross-flank retrieval ablation (foreground-only, macro metrics).
Flip augmentation does not improve cross-flank mAP and reduces within-flank
accuracy by 12\%.}
\label{tab:crossflank}
\begin{tabular}{@{}lcccccc@{}}
\toprule
& \multicolumn{2}{c}{Within-flank} & \multicolumn{2}{c}{Cross-flank} & Cross/Within & Discrim- \\
\cmidrule(lr){2-3}\cmidrule(lr){4-5}
Model & mAP & CMC@1 & mAP & CMC@1 & ratio & inability \\
\midrule
ArcFace+MiewID (no flip) & \textbf{0.716} & \textbf{77.5\%} & 0.477 & 53.9\% & 0.666 & 0.417 \\
ArcFace+MiewID (flip aug) & 0.632 & 70.7\% & \textbf{0.490} & \textbf{55.3\%} & 0.774 & \textbf{0.427} \\
\midrule
$\Delta$ (flip $-$ no flip) & $-$0.084 & $-$6.8pp & +0.013 & +1.4pp & +0.108 & +0.010 \\
\bottomrule
\end{tabular}
\end{table}

Cross-flank macro mAP is nearly identical between models (0.477 vs.\ 0.490,
$\Delta{=}+0.013$), while within-flank mAP drops from 0.716 to 0.632 with
flips.  The higher cross/within ratio for the flip model (0.774 vs.\ 0.666) is
driven entirely by its lower within-flank performance, not by improved
cross-flank matching.  Both models achieve discriminability above 0.4 (same-ID
cross-flank similarity ${\approx}0.47$ vs.\ different-ID ${\approx}0.04$),
confirming that the model learns cross-viewpoint generalisation from identity
supervision alone.  Due to this, we omit flips and evaluate cross-flank
performance as the direct test of laterality robustness.

% ------------------------------------------------------------------
\subsection{Lorentz Hyperbolic Embeddings}\label{sec:lorentz_results}

For the main benchmark, we keep the matched objective-pair families in one
table and read the results in two layers: descriptive ranking first, paired
claims second.

\begin{table}[t]
\centering
\caption{TEST-only retrieval summary for the matched objective-pair benchmark.
Values are family-level mean $\pm$ std over seeds 42/43/44.  Full-RGB and cross-flank endpoints both report macro mAP and macro
CMC@1/5/10.  These rows describe benchmark ordering; pairwise analyses appear on the Appendix's Wilcoxon tables.}
\label{tab:lorentz_test}
\resizebox{\textwidth}{!}{%
\begin{tabular}{@{}lcccccccc@{}}
\toprule
& \multicolumn{4}{c}{Full RGB} & \multicolumn{4}{c}{Cross-Flank} \\
\cmidrule(lr){2-5}\cmidrule(lr){6-9}
Model & mAP & CMC@1 & CMC@5 & CMC@10 & mAP & CMC@1 & CMC@5 & CMC@10 \\
\midrule
ArcFace O0 (MSv3) & 0.483 $\pm$ 0.011 & 74.3\% $\pm$ 0.9\% & 80.7\% $\pm$ 1.0\% & 83.8\% $\pm$ 0.4\% & 0.576 $\pm$ 0.049 & 60.7\% $\pm$ 4.9\% & 69.0\% $\pm$ 4.3\% & 75.1\% $\pm$ 2.6\% \\
ArcFace O1 (MSv3) & 0.469 $\pm$ 0.010 & 74.3\% $\pm$ 0.9\% & 80.4\% $\pm$ 1.5\% & 82.4\% $\pm$ 1.0\% & 0.566 $\pm$ 0.026 & 62.4\% $\pm$ 1.8\% & 69.9\% $\pm$ 4.7\% & 76.2\% $\pm$ 5.6\% \\
Lorentz O0 (MSv3) & 0.548 $\pm$ 0.012 & 76.3\% $\pm$ 1.5\% & 83.1\% $\pm$ 1.0\% & 86.3\% $\pm$ 0.7\% & 0.689 $\pm$ 0.032 & 72.0\% $\pm$ 4.7\% & 81.5\% $\pm$ 0.7\% & 84.8\% $\pm$ 2.5\% \\
Lorentz O1 (MSv2) & 0.515 $\pm$ 0.014 & 75.0\% $\pm$ 1.7\% & 83.5\% $\pm$ 0.3\% & \textbf{87.6\% $\pm$ 0.9\%} & 0.653 $\pm$ 0.011 & 71.2\% $\pm$ 4.9\% & \textbf{84.4\% $\pm$ 1.2\%} & \textbf{88.8\% $\pm$ 1.0\%} \\
Lorentz O1 (MSv3) & \textbf{0.557 $\pm$ 0.009} & \textbf{77.3\% $\pm$ 1.7\%} & \textbf{84.3\% $\pm$ 1.4\%} & 87.1\% $\pm$ 1.4\% & \textbf{0.692 $\pm$ 0.010} & 72.5\% $\pm$ 2.2\% & 81.9\% $\pm$ 3.4\% & 87.0\% $\pm$ 2.9\% \\
Triplet T0 (MSv3) & 0.510 $\pm$ 0.002 & 76.7\% $\pm$ 1.1\% & 81.6\% $\pm$ 0.2\% & 83.8\% $\pm$ 1.2\% & 0.681 $\pm$ 0.013 & \textbf{73.0\% $\pm$ 2.3\%} & 81.3\% $\pm$ 2.4\% & 84.1\% $\pm$ 1.2\% \\
Triplet T1 (MSv3) & 0.479 $\pm$ 0.004 & 76.6\% $\pm$ 1.1\% & 82.6\% $\pm$ 0.8\% & 85.1\% $\pm$ 0.8\% & 0.647 $\pm$ 0.003 & 71.5\% $\pm$ 0.4\% & 78.6\% $\pm$ 0.7\% & 83.2\% $\pm$ 0.7\% 
\\
\bottomrule
\end{tabular}}
\end{table}

\Cref{tab:lorentz_test} is the cleanest example of why we split the narrative.
Descriptively, Lorentz O1 (MSv3) is the strongest full-image retrieval model on
macro mAP (0.557 $\pm$ 0.009) and also the strongest cross-flank model on macro
mAP (0.692 $\pm$ 0.010). The matched
Lorentz O0 vs.\ O1 AP tests are not supported on full-RGB AP, cross-flank AP,
or any CMC endpoint.

Axis~2 adds the second half of the story.  Under the active rank-primary
calibration rule, the cross-space conclusion is unchanged: laterality claims
should be read through mirror-rank percentile, not raw mirror values alone.
In that view, Triplet T1 is the most pattern-aware laterality model, while
Lorentz O1 (MSv3) remains the strongest retrieval model.  Those findings tell us that the leaderboard and the laterality
diagnostic reward different kinds of success.  The appendix tables keep both
views visible rather than forcing them into one number
(\cref{tab:supp_crossflank_finalists,tab:supp_axis2_mitigation}).

% =====================================================================
% 5  DISCUSSION  (~1.0 page)
% =====================================================================
\section{Discussion}\label{sec:discussion}
% Target: ~0.3 pages

Our study covers one species in one ecosystem, so broader validation across
taxa and sites remains open.  The statistical takeaway from the active bundle
is narrower than a casual leaderboard read.  Once we insist on matched
objective pairs, complete seeds, Fisher combination, and Holm correction, most
apparent gaps stop being claimable pairwise wins.

\begin{figure}[!htbp]
\centering
\begin{subfigure}[t]{\textwidth}
  \centering
  \includegraphics[width=0.82\textwidth,trim=155 20 18 35,clip]{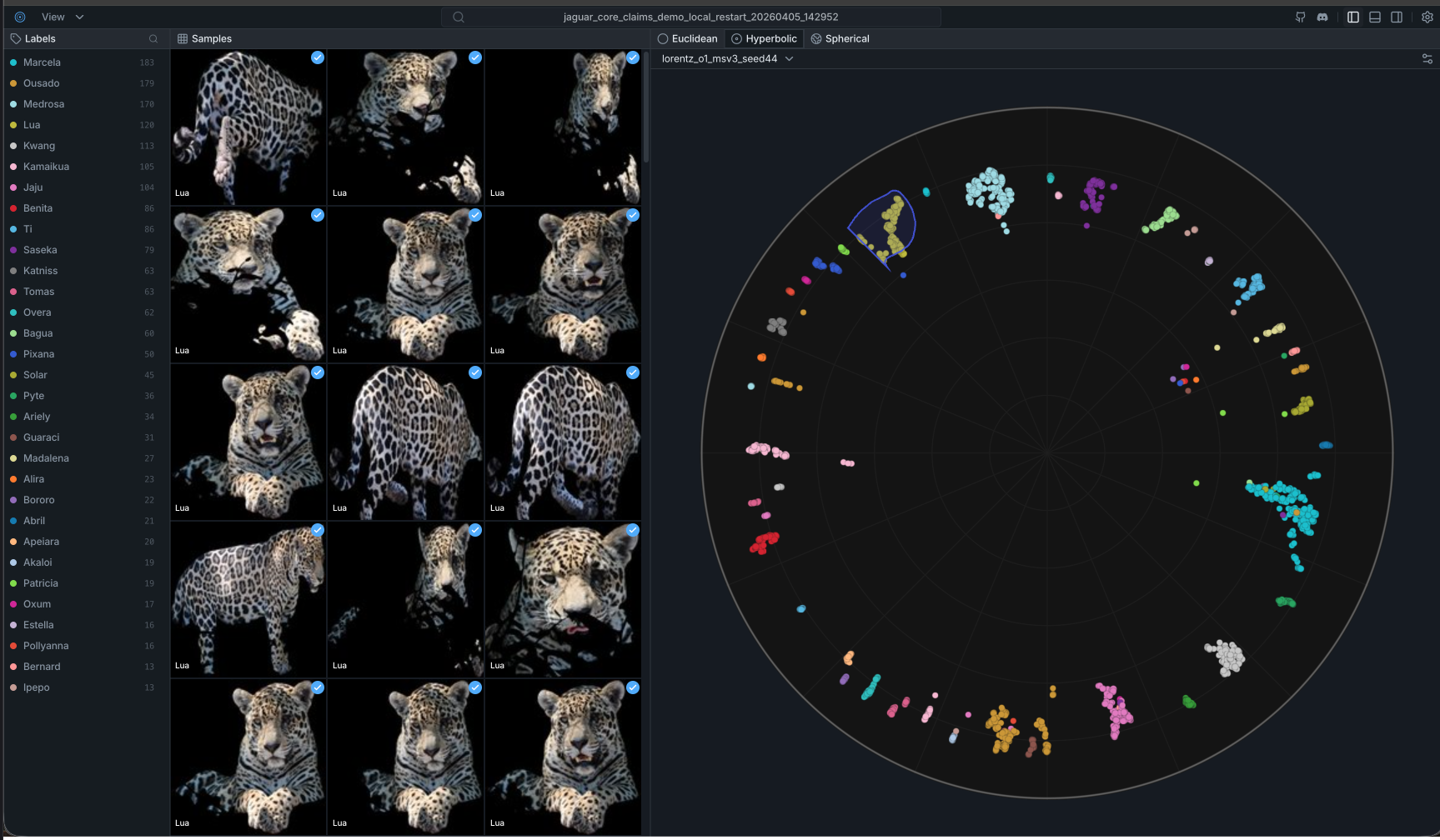}
  \caption{Lorentz O1 in the Poincar\'e disk; points nearer the center tend to
  be more ambiguous or lower quality (e.g. low mask solidity).}
  \label{fig:hyperview_hyperbolic}
\end{subfigure}
\vspace{0.35em}
\begin{subfigure}[t]{\textwidth}
  \centering
  \includegraphics[width=0.82\textwidth,trim=155 20 18 35,clip]{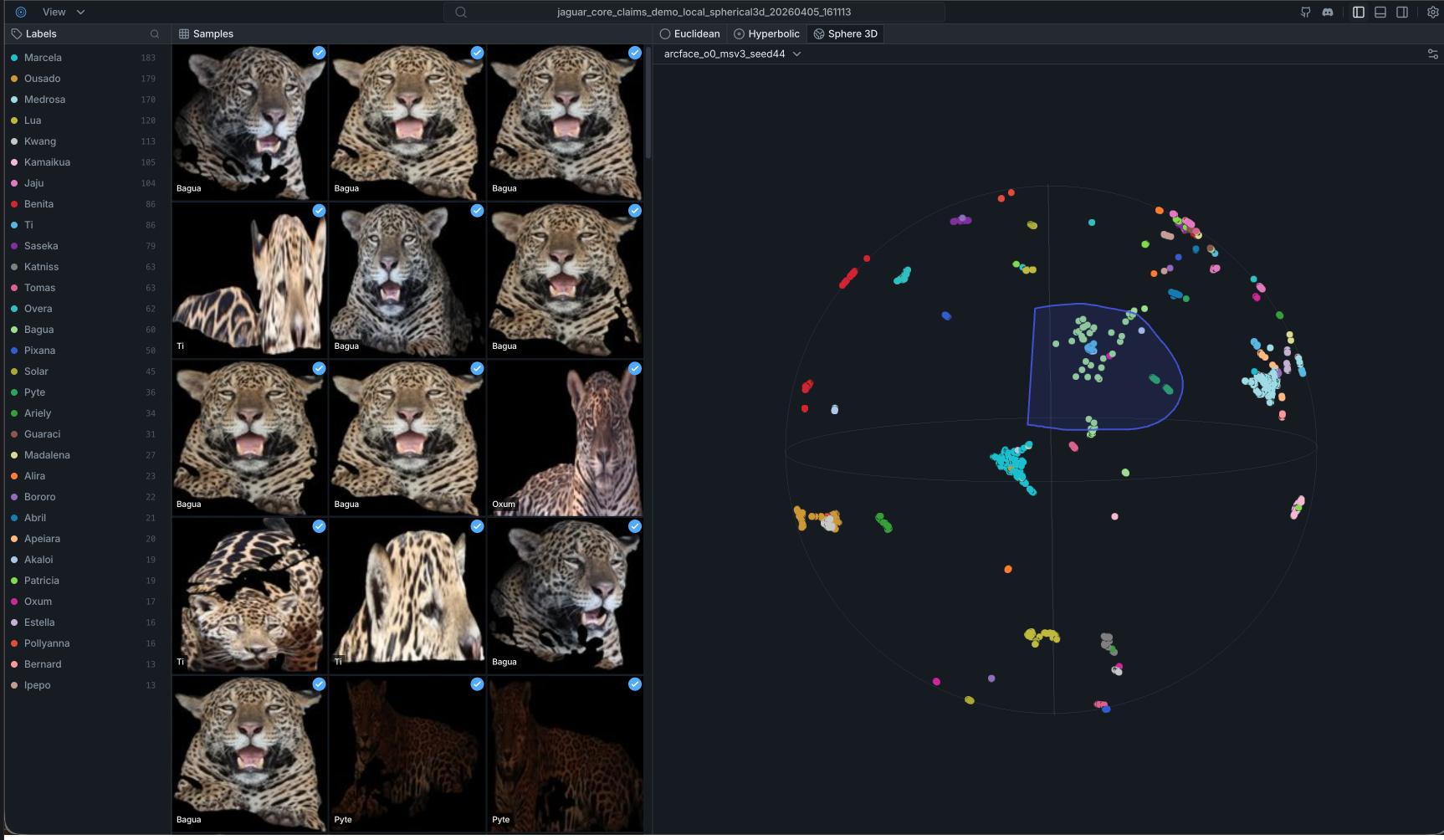}
  \caption{ArcFace O0 in hyperspherical projection; margin penalization appears
  as sharper class boundaries and stronger isolation.}
  \label{fig:hyperview_spherical}
\end{subfigure}
\caption{\textbf{Qualitative embedding inspection with UMAP~\cite{mcinnes2018umap} in HyperView.}  The interface
shows image thumbnails next to the latent arrangement and includes a lasso tool
for selecting regions for qualitative inspection, near-duplicate discovery, and
difficult re-identification cases. Hard cases produce
overlapping embedding representations.  These panels are qualitative
explorations and part of the iterative model-guided dataset curation process.}
\label{fig:hyperview_embeddings}
\end{figure}

The cross-flank ablation (\cref{sec:crossflank}) clarifies the laterality
question: the relevant target is cross-flank retrieval, not mirror similarity
in isolation.  Flip augmentation does not improve cross-flank re-ID and
reduces within-flank accuracy, so the right augmentation strategy is to omit flips and to
measure cross-flank performance directly.  More broadly, we advocate for 
two layers of reporting: descriptive benchmarks to show where models land,
and paired endpoint-specific tests to show which differences survive correction.
The qualitative inspection of the embedding space in multiple geometries
(Euclidean and Lorentz) gives a more direct sense of which images are cleanly
separated and which ones remain ambiguous.  We use the
HyperView tool~\cite{hyperview2025} for this.  By showing image thumbnails
alongside the latent layout, \Cref{fig:hyperview_embeddings} gives an
intuitive view of representation quality that complements the scalar retrieval
metrics.

\clearpage
% =====================================================================
% 6  CONCLUSION  (~0.5 pages)
% =====================================================================
\section{Conclusion}\label{sec:conclusion}
% Target: ~0.5 pages

We have presented a diagnostic evaluation protocol for shortcut learning in
jaguar re-identification.  The mask-annotated benchmark makes it possible to
measure context reliance and laterality collapse directly, and the mitigation
families in this paper are included as case studies under that common lens.
This framing matters because high retrieval accuracy can still come from the
wrong evidence.

Across frozen baselines and mitigation case studies, shortcut learning is evident and variable.
Some strong encoders still extract substantial identity
signal from background context, and laterality sensitivity does not track
context reliance in a simple way.  In our current benchmarks, the MiewID-MSv3 model appears as the strongest backbone and baseline, and appears improved by anti-symmetry regularization and fine-tuning on hyperbolic space, but the paired tests are more conservative:
sparse and specific to test conditions. As a takeaway: good jaguar retrieval scores are not enough on their own, inspection of the query results by experts is also necessary.

We encourage the wildlife re-ID community to report the background reliance 
and laterality diagnostics alongside mAP and CMC, and to add paired
significance tests whenever matched objective pairs are available.  Without
probing \emph{what} a model relies on, accuracy alone cannot tell us whether a
re-ID system is trustworthy for the conservation decisions that depend on it.

\clearpage
\appendix
These appendices collect the material that previously sat in the supplement.
The goal is practical: keep the arXiv version self-contained without splitting
the paper across two PDFs.  We include architecture summaries, training and
selection protocol notes, benchmark breakdowns, laterality analyses, and the
paired Wilcoxon summaries used for claim-facing language in the main text.

\paragraph{Data and artifact availability.}
Public resources for this paper are available at the Jaguar ID Project website
(\href{https://www.jaguaridproject.com/}{Jaguar ID Project website}), the
Hugging Face dataset page
(\href{https://huggingface.co/datasets/jaguaridentification/jaguars}{Hugging Face dataset page}),
and the two Kaggle challenges:
\href{https://www.kaggle.com/competitions/jaguar-re-id}{Round 1 Kaggle challenge}
and
\href{https://www.kaggle.com/competitions/round-2-jaguar-reidentification-challenge}{Round 2 Kaggle challenge}.
The claim-facing numbers in this arXiv version come from the local canonical
bundle registered in the repository.

% =====================================================================
\section*{Backbone Architecture Summary}

\Cref{tab:supp_backbones} lists all 16 backbone architectures evaluated in
this study.  Sixteen are included in the frozen baseline evaluation
(Table~1 in the main paper); eight are additionally fine-tuned under one or
more training objectives.  Two backbones (ConvNeXt-Tiny, Swin-Tiny) appear
only in the fine-tuning benchmarks.  Together, these 16 backbones yield the
102 model checkpoints discussed throughout.

\begin{table}[t]
\centering
\caption{All 16 backbone architectures evaluated.  \emph{Dim}\,=\,frozen
embedding dimension before any projection head.
\emph{Frozen}\,=\,included in the frozen baseline evaluation
(Table~1 in the main paper).
Fine-tuning abbreviations:
A\,=\,ArcFace, T\,=\,Triplet, L\,=\,Lorentz, AS\,=\,Anti-symmetry.}
\label{tab:supp_backbones}
\resizebox{\textwidth}{!}{%
\begin{tabular}{@{}lllrcl@{}}
\toprule
Backbone & Architecture & Pre-training & Dim & Frozen & Fine-tuned \\
\midrule
\multicolumn{6}{@{}l}{\emph{Wildlife-specific pre-training}} \\[2pt]
MiewID-MSv2~\cite{otarashvili2024multispecies}       & EfficientNetV2-M  & Wildlife (49\,spp.)    & 2152 & $\checkmark$           & A, T, L, AS \\
MiewID-MSv3~\cite{otarashvili2024multispecies}       & SwinV2            & Wildlife               & 2152 & $\checkmark$           & A, T, L     \\
MegaDescriptor-L~\cite{vcermak2024wildlifedatasets}  & Swin-L            & Wildlife               & 1536 & $\checkmark$           & A, T        \\
\midrule
\multicolumn{6}{@{}l}{\emph{Self-supervised / multi-task}} \\[2pt]
DINOv3-ViT~\cite{simeoni2025dinov3}         & ViT-S/16   & Self-sup.\ (LVD-142M)  &  384 & $\checkmark^{\dagger}$ & A           \\
DINOv3-ConvNeXt~\cite{simeoni2025dinov3}    & ConvNeXt-S & Self-sup.\ (LVD-142M)  &  768 & $\checkmark^{\dagger}$ & ---         \\
EVA-02~\cite{fang2024eva}                   & ViT-L/14   & CLIP + MIM             & 1024 & $\checkmark$           & ---         \\
C-RADIO-v4~\cite{ranzinger2026cradio}       & ViT-H (CPE) & Multi-teacher distill. & 2560 & $\checkmark$          & ---         \\
I-JEPA~\cite{assran2023ijepa}               & ViT-H/14   & Predictive SSL (IN-1K) & 1280 & $\checkmark$           & ---         \\
\midrule
\multicolumn{6}{@{}l}{\emph{ImageNet-supervised}} \\[2pt]
EfficientNetV2-M~\cite{tan2021efficientnetv2} & CNN      & Supervised (IN-1K)     & 1280 & $\checkmark$           & A, T        \\
ConvNeXtV2-Base~\cite{woo2023convnextv2}     & CNN       & FCMAE + FT (IN-22K)    & 1024 & $\checkmark$           & ---         \\
ConvNeXt-Tiny~\cite{liu2022convnext}         & CNN       & Supervised (IN-1K)     &  768 & ---                    & A, T        \\
Swin-Tiny~\cite{liu2021swin}                & Swin-T     & Supervised (IN-1K)     &  768 & ---                    & A, T        \\
ResNet50~\cite{he2016deep}                  & CNN        & Supervised (IN-1K V2)  & 2048 & $\checkmark$           & ---         \\
ResNet152~\cite{he2016deep}                 & CNN        & Supervised (IN-1K V2)  & 2048 & $\checkmark$           & ---         \\
\midrule
\multicolumn{6}{@{}l}{\emph{Domain-specific (face)}} \\[2pt]
meFeM-B~\cite{borets2026mefem}              & ViT-B/16   & Face matching          &  768 & $\checkmark$           & ---         \\
\midrule
\multicolumn{6}{@{}l}{\emph{Classical computer vision}} \\[2pt]
HotSpotter~\cite{crall2013hotspotter}       & SIFT + VLAD & None (hand-crafted)   & 8192 & $\checkmark$           & ---         \\
\bottomrule
\end{tabular}}
\smallskip\\
{\footnotesize $^\dagger$Evaluated with both CLS-token and mean-pooled aggregation, yielding 4 rows in the frozen baseline table.}
\end{table}

\section*{Validation Coreset Construction}

All hyperparameters are tuned and checkpoints are selected using a fixed
held-out validation coreset drawn from the 1{,}895 labeled training images.
We select 355 images (18.7\%) via an embedding-aware Facility Location
coverage objective computed on L2-normalised 2{,}152-d embeddings from the
best ArcFace+MiewID-MSv2 checkpoint (seed 42), with cosine similarity as the
similarity measure.  Selection proceeds in three phases: (1) per-identity
floor selection with $|S_i| \ge \max(3,\lceil 0.15\,|V_i| \rceil)$,
(2) global greedy fill to a target coverage ratio of 0.95, and
(3) boundary enrichment via within-identity swaps that prioritise images with
high cross-identity similarity (hard examples).  After boundary enrichment,
the final coreset achieves a coverage ratio of 0.9457 and contains 27.9\%
boundary images.  The coreset covers all 31 identities with a minimum of 3
images per identity.

\section*{Comparison Protocol Harmonization}

To make the main-paper comparisons directly comparable, we state the full
training and selection protocol for every trainable family in one place.
All main-paper reruns use the same fixed validation coreset and are reported
at the family level as test mean $\pm$ std over seeds 42/43/44.  Single
checkpoints are kept only as representative artifacts for figures, audits,
and qualitative examples.

\begin{table}[t]
\centering
\caption{Explicit comparison protocol for the trainable families referenced in
the main paper and mitigation supplement.  ``FG cutouts'' means the alpha-mask
foreground cutout is applied at load time.  Lorentz family selection is first
done at the family level by mean validation ib-mAP across seeds; the
representative checkpoint is then chosen on validation statistics only.}
\label{tab:supp_protocol_harmonization}
\resizebox{\textwidth}{!}{%
\begin{tabular}{@{}lllllll@{}}
\toprule
Family / model & Training input & Size & Checkpoint selection & Seed handling & Final statistic & Representative checkpoint rule \\
\midrule
ArcFace+MiewID & FG cutouts & 440 & Lowest val loss on coreset & Seeds 42/43/44 & Test mean $\pm$ std & Lowest val-loss seed \\
ArcFace+MiewID+AntiSym ($\tau{=}0.3,\lambda{=}0.5$) & FG cutouts & 440 & Lowest val loss on coreset & Seeds 42/43/44; hyperparams fixed before reruns & Test mean $\pm$ std & Lowest val-loss seed \\
Lorentz+MSv3+O0 & FG cutouts + solidity & 320 & Family: mean val ib-mAP; seed ckpt: val loss & Seeds 42/43/44 & Test mean $\pm$ std & Highest val ib-mAP, lower val loss tiebreak \\
Lorentz+MSv2+O1 & FG cutouts + solidity & 320 & Family: mean val ib-mAP; seed ckpt: val loss & Seeds 42/43/44 & Test mean $\pm$ std & Highest val ib-mAP, lower val loss tiebreak \\
Lorentz+MSv3+O1 & FG cutouts + solidity & 320 & Family: mean val ib-mAP; seed ckpt: val loss & Seeds 42/43/44 & Test mean $\pm$ std & Highest val ib-mAP, lower val loss tiebreak \\
ArcFace+DINOv3 & FG cutouts & 224 & Lowest val loss on coreset & Seeds 42/43/44 & Test mean $\pm$ std & Lowest val-loss seed \\
Triplet benchmark family & FG cutouts & Backbone-dependent & Highest val ib-mAP & Three seeds per cell & Test mean $\pm$ std & Highest val ib-mAP seed \\
\bottomrule
\end{tabular}}
\end{table}

\begin{table}[t]
\centering
\scriptsize
\caption{Per-seed TEST metrics for the six matched objective-pair families in
the main benchmark table (seeds 42/43/44).  We report full-RGB macro mAP,
full-RGB macro CMC@1, cross-flank macro mAP, cross-flank macro CMC@1, and raw
Axis~2 mirror similarity, together with the checkpoint retained for
traceability.}
\label{tab:supp_main_per_seed}
\resizebox{\textwidth}{!}{%
\begin{tabular}{@{}lllccccc@{}}
\toprule
Family & Seed & Checkpoint & Full-RGB mAP & Full-RGB CMC@1 & Cross-Flank mAP & Cross-Flank CMC@1 & Raw mirror \\
\midrule
ArcFace O0 (MSv3) & 42 & \texttt{arcface\_miewid\_msv3\_O0\_seed42\_best.pth} & 0.475 & 73.2\% & 0.582 & 65.1\% & 0.858 \\
ArcFace O0 (MSv3) & 43 & \texttt{arcface\_miewid\_msv3\_O0\_seed43\_best.pth} & 0.476 & 75.5\% & 0.513 & 53.8\% & 0.850 \\
ArcFace O0 (MSv3) & 44 & \texttt{arcface\_miewid\_msv3\_O0\_seed44\_best.pth} & 0.499 & 74.2\% & 0.633 & 63.1\% & 0.857 \\
ArcFace O1 (MSv3) & 42 & \texttt{arcface\_miewid\_msv3\_O1\_seed42\_best.pth} & 0.479 & 75.4\% & 0.576 & 64.8\% & 0.849 \\
ArcFace O1 (MSv3) & 43 & \texttt{arcface\_miewid\_msv3\_O1\_seed43\_best.pth} & 0.473 & 74.4\% & 0.530 & 60.4\% & 0.844 \\
ArcFace O1 (MSv3) & 44 & \texttt{arcface\_miewid\_msv3\_O1\_seed44\_best.pth} & 0.456 & 73.1\% & 0.591 & 61.9\% & 0.848 \\
Lorentz O0 (MSv3) & 42 & \texttt{lorentz\_miewid\_msv3\_O0\_seed42\_best.pth} & 0.531 & 74.3\% & 0.662 & 66.7\% & 0.290 \\
Lorentz O0 (MSv3) & 43 & \texttt{lorentz\_miewid\_msv3\_O0\_seed43\_best.pth} & 0.556 & 77.6\% & 0.670 & 71.1\% & 0.287 \\
Lorentz O0 (MSv3) & 44 & \texttt{lorentz\_miewid\_msv3\_O0\_seed44\_best.pth} & 0.557 & 77.2\% & 0.734 & 78.1\% & 0.292 \\
Lorentz O1 (MSv2) & 42 & \texttt{lorentz\_miewid\_msv2\_O1\_seed42\_best.pth} & 0.500 & 72.7\% & 0.668 & 76.9\% & 0.288 \\
Lorentz O1 (MSv2) & 43 & \texttt{lorentz\_miewid\_msv2\_O1\_seed43\_best.pth} & 0.512 & 75.7\% & 0.643 & 64.9\% & 0.291 \\
Lorentz O1 (MSv2) & 44 & \texttt{lorentz\_miewid\_msv2\_O1\_seed44\_best.pth} & 0.533 & 76.7\% & 0.647 & 71.7\% & 0.293 \\
Lorentz O1 (MSv3) & 42 & \texttt{lorentz\_miewid\_msv3\_O1\_seed42\_best.pth} & 0.555 & 75.5\% & 0.697 & 70.9\% & 0.288 \\
Lorentz O1 (MSv3) & 43 & \texttt{lorentz\_miewid\_msv3\_O1\_seed43\_best.pth} & 0.568 & 79.6\% & 0.679 & 71.0\% & 0.292 \\
Lorentz O1 (MSv3) & 44 & \texttt{lorentz\_miewid\_msv3\_O1\_seed44\_best.pth} & 0.547 & 76.7\% & 0.701 & 75.6\% & 0.306 \\
Triplet T0 (MSv3) & 42 & \texttt{triplet\_miewid\_msv3\_T0\_seed42\_best.pth} & 0.510 & 77.7\% & 0.663 & 70.0\% & 0.738 \\
Triplet T0 (MSv3) & 43 & \texttt{triplet\_miewid\_msv3\_T0\_seed43\_best.pth} & 0.512 & 75.3\% & 0.696 & 75.5\% & 0.743 \\
Triplet T0 (MSv3) & 44 & \texttt{triplet\_miewid\_msv3\_T0\_seed44\_best.pth} & 0.508 & 77.2\% & 0.683 & 73.5\% & 0.741 \\
Triplet T1 (MSv3) & 42 & \texttt{triplet\_miewid\_msv3\_T1\_seed42\_best.pth} & 0.476 & 75.5\% & 0.651 & 72.1\% & 0.434 \\
Triplet T1 (MSv3) & 43 & \texttt{triplet\_miewid\_msv3\_T1\_seed43\_best.pth} & 0.477 & 76.1\% & 0.647 & 71.0\% & 0.433 \\
Triplet T1 (MSv3) & 44 & \texttt{triplet\_miewid\_msv3\_T1\_seed44\_best.pth} & 0.484 & 78.1\% & 0.643 & 71.4\% & 0.438 
\\
\bottomrule
\end{tabular}}
\end{table}

%\paragraph{Reproducibility note.}
%\Cref{tab:lorentz_test} and \cref{tab:supp_main_per_seed} are generated from
%the active v5 canonical bundle.  The corresponding artifact pack includes the
%seed-level CSVs, family summaries, claim statements, and provenance files used
%to audit the paper-facing tables.

% =====================================================================
\section*{Segmentation Masks and Solidity}\label{sec:supp_masks}

\paragraph{Mask generation (SAM~3).}
We generate a jaguar foreground mask for each image using SAM~3~\cite{carion2025sam}
with the text prompt \texttt{"jaguar"}.  The dataset used in our experiments
stores this binary mask as the alpha channel of the RGBA PNG; all experiments
use these masks, so reproducing our image variants does not require re-running
SAM.

\begin{figure}[t]
\centering
\includegraphics[width=\textwidth]{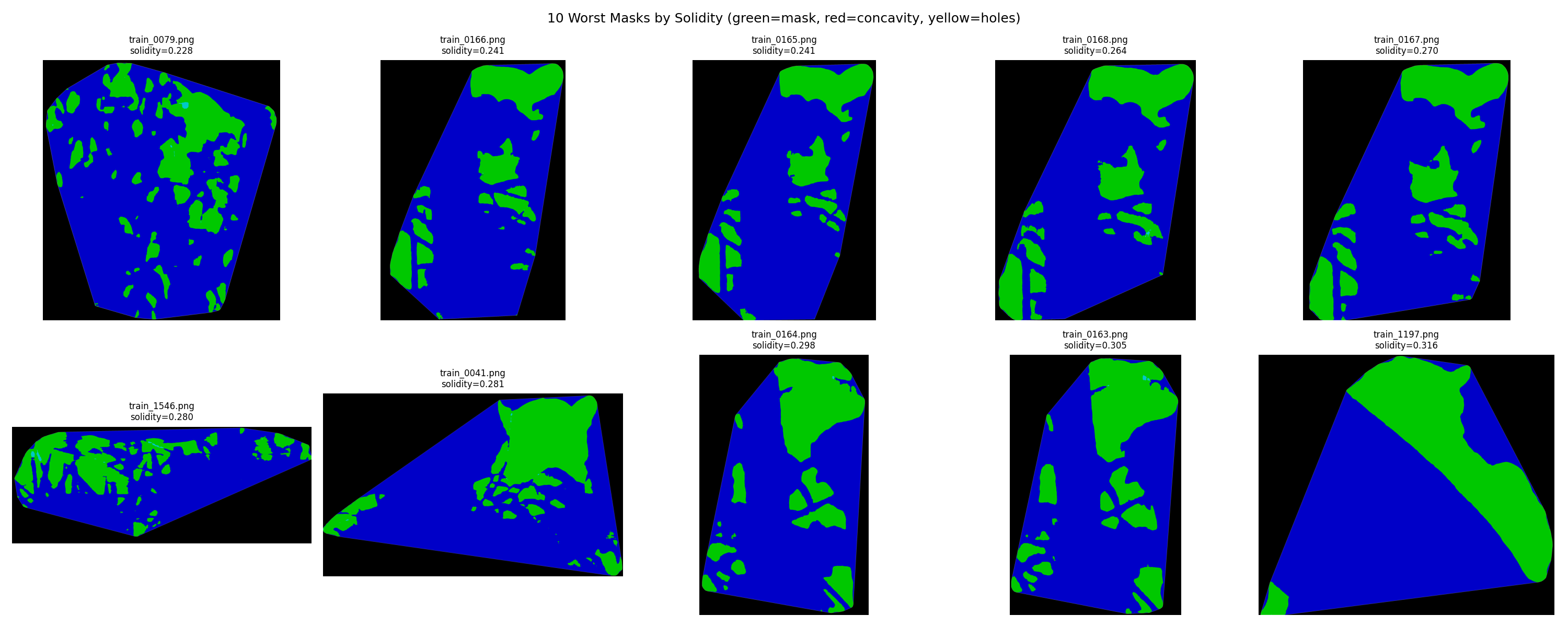}
\caption{\textbf{Lowest-solidity masks.}  Examples from the bottom of the solidity
distribution.  The most common artifact we see is partial occlusion (often vegetation),
which removes parts of the jaguar or introduces small holes.}
\label{fig:supp_worst_masks}
\end{figure}

\paragraph{Quality control and typical failure modes.}
We compute per-image mask diagnostics (including solidity; Eq.~\ref{eq:solidity_def})
for the full train and test splits and manually inspect low-solidity examples.
In these cases, the most common artifact is partial occlusion by small vegetation,
which can erase thin body parts or introduce small boundary holes.  In our
inspection we mostly see missing foreground, rather than large background
regions incorrectly included as foreground (\cref{fig:supp_worst_masks}).
Only 0.0770 of training masks and 0.0701 of test masks have $s < 0.5$
(\cref{tab:supp_solidity_stats}).

\paragraph{Solidity definition (exact).}
Let $M$ be a binary foreground mask and let $H(M)$ denote the convex hull of
its foreground pixels.  We define mask solidity:
\begin{equation}\label{eq:solidity_def}
  s(M) = \frac{|M|}{|H(M)|}\,.
\end{equation}
Lower $s$ indicates a less compact mask under this convex-hull proxy (often due
to occlusion, partial crops, or fragmentation).  In Lorentz training we use
$s$ as a continuous quality signal in the variance mapping $\sigma^2(s)$
(Sec.~\ref{sec:supp_lorentz_loss}).

\begin{table}[t]
\centering
\caption{Mask solidity statistics computed from the SAM~3 alpha masks used in our experiments
(Eq.~\ref{eq:solidity_def}).  ``Frac($s<0.5$)'' is the fraction of images with
solidity below 0.5.}
\label{tab:supp_solidity_stats}
\begin{tabular}{@{}lrrrrrr@{}}
\toprule
Split & $n$ & Mean & Median & Min & Max & Frac($s<0.5$) \\
\midrule
Train & 1{,}895 & 0.6937 & 0.7094 & 0.2277 & 0.9678 & 0.0770 \\
Test  &   371 & 0.6883 & 0.6967 & 0.2388 & 0.9561 & 0.0701 \\
\bottomrule
\end{tabular}
\end{table}

\begin{figure}[t]
\centering
\includegraphics[width=0.82\textwidth]{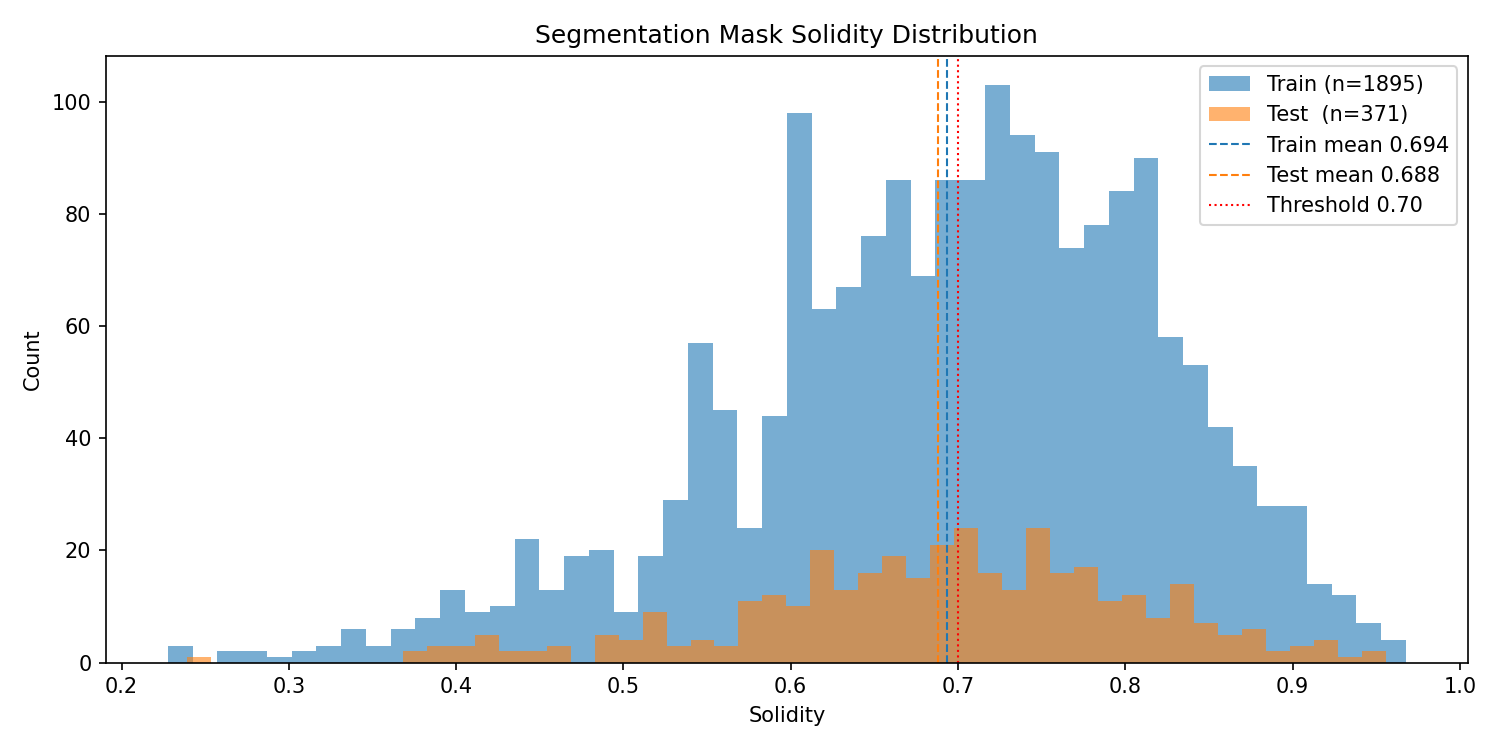}
\caption{\textbf{Mask solidity distribution} for the train ($n{=}1{,}895$) and
test ($n{=}371$) splits, computed from SAM~3 alpha masks (Eq.~\ref{eq:solidity_def}).}
\label{fig:supp_solidity_dist}
\end{figure}

\section{Leakage-Controlled Background-Only Variant via Inpainting}\label{sec:supp_inpainting}

\paragraph{Motivation.}
The companion ratio \BGSFG{} uses \texttt{background\_only} images constructed
by zeroing out the jaguar pixels using the alpha mask.  This introduces a
jaguar-shaped silhouette hole and boundary artifacts, so \BGSFG{} should be
interpreted as a \emph{context} ratio (background plus silhouette-hole /
boundary cues) rather than a purely background-only ratio.

\paragraph{Inpainted background-only images.}
To measure background reliance without silhouette leakage, we generate a
leakage-controlled variant, \texttt{background\_inpainted}, where the jaguar
region is removed and inpainted to plausible background content using
FLUX.1-Fill-dev~\cite{blackforestlabs2024flux}.
The inpainting prompt is
\begin{quote}
\small\ttfamily
"lush tropical jungle vegetation, dense green leaves,\\
ferns, vines, and undergrowth, natural forest floor,\\
no animals, no creatures, no wildlife, only plants"
\end{quote}
We produce three generations per test image with different random seeds
to test whether the BG/FG ratio is sensitive to variation in the
inpainted content.

\paragraph{Reproducibility settings.}
Unless stated otherwise, we use:
\begin{quote}
\small\ttfamily
num\_inference\_steps=50, guidance\_scale=30.0,\\
max\_size=1024, seed=0, convex\_hull=true,\\
mask\_dilate\_px=0,\\
prompt="lush tropical jungle vegetation, dense green leaves,\\
ferns, vines, and undergrowth, natural forest floor,\\
no animals, no creatures, no wildlife, only plants"
\end{quote}
Images are resized so the longest edge is at most 1024 pixels, preserving
aspect ratio, and dimensions are rounded to multiples of 8 for diffusion
compatibility.

\paragraph{Sensitivity protocol (seeds, prompts, non-generative baseline).}
To test robustness beyond a single inpainting setting, we run the evaluation three times, with different random seeds, for the inpainting generation. 

\paragraph{Leakage-controlled \BGFG{}.}
The inpainted \texttt{background\_inpainted} variant replaces the jaguar
foreground with a generative fill (FLUX.1-Fill-dev~\cite{blackforestlabs2024flux}),
preserving the natural background without a silhouette-shaped hole.
The leakage-controlled ratio is:
\begin{equation}\label{eq:bgfg_inpainted_supp}
  \BGFG = \frac{\text{mAP}(\texttt{background\_inpainted})}
               {\text{mAP}(\texttt{foreground\_only})}\,.
\end{equation}
\Cref{tab:supp_bgfg_inpainted} reports inpainted \BGFG{} for frozen baselines
and mitigation finalists.  Values are lower than \BGSFG{} for most
models, because the silhouette-shaped hole in the na\"ive
\texttt{background\_only} construction leaks shape information.

\begin{table}[t]
\centering
\caption{Leakage-controlled inpainted \BGFG{} results across frozen baselines
and mitigation finalists (test split).  Full~mAP, FG~mAP, and Inpainted~mAP are
identity-balanced (macro), computed on \texttt{full\_rgb},
\texttt{foreground\_only}, and \texttt{background\_inpainted}, respectively.
Models are grouped by type and sorted by \BGFG{}.}
\label{tab:supp_bgfg_inpainted}
\resizebox{\textwidth}{!}{%
\begin{tabular}{@{}lcccccc@{}}
\toprule
Model & Full mAP & FG mAP & Inpainted mAP & \BGFG{} & \BGSFG{} & Risk \\
\midrule
\multicolumn{7}{l}{\emph{Frozen baselines}} \\
MiewID-MSv2             & 0.297 & 0.292 & 0.151 & 0.516 & 0.848 & LOW \\
HotSpotter+VLAD         & 0.288 & 0.271 & 0.145 & 0.535 & 0.846 & LOW \\
MiewID-MSv3             & 0.303 & 0.304 & 0.178 & 0.586 & 0.869 & LOW \\
EVA-02                  & 0.281 & 0.303 & 0.200 & 0.661 & 0.940 & LOW \\
DINOv3-ViT CLS          & 0.298 & 0.296 & 0.228 & 0.768 & 1.020 & LOW \\
MegaDescriptor          & 0.249 & 0.286 & 0.220 & 0.769 & 1.020 & LOW \\
C-RADIO-v4              & 0.255 & 0.253 & 0.198 & 0.784 & 1.221 & LOW \\
DINOv3-ViT Pooled       & 0.277 & 0.277 & 0.228 & 0.821 & 1.179 & LOW \\
ResNet50                & 0.218 & 0.227 & 0.205 & 0.905 & 1.125 & LOW \\
DINOv3-ConvNeXt CLS     & 0.208 & 0.211 & 0.191 & 0.907 & 1.161 & LOW \\
DINOv3-ConvNeXt Pooled  & 0.209 & 0.208 & 0.191 & 0.920 & 1.355 & LOW \\
ResNet152               & 0.218 & 0.227 & 0.214 & 0.942 & 1.125 & LOW \\
ConvNeXtV2-Base         & 0.210 & 0.209 & 0.209 & 0.997 & 1.058 & MEDIUM \\
EfficientNetV2-M        & 0.200 & 0.199 & 0.206 & 1.034 & 1.009 & MEDIUM \\
meFeM-B                 & 0.186 & 0.198 & 0.223 & 1.125 & 1.058 & HIGH \\
I-JEPA                  & 0.238 & 0.238 & 0.278 & 1.166 & 1.271 & HIGH \\
\midrule
\multicolumn{7}{l}{\emph{Lorentz finalists}} \\
Lorentz+MSv3+O0         & 0.598 & 0.637 & 0.105 & 0.165 & 0.245 & LOW \\
Lorentz+MSv2+O1         & 0.609 & 0.658 & 0.128 & 0.195 & 0.301 & LOW \\
Lorentz+MSv3+O1         & 0.629 & 0.669 & 0.129 & 0.194 & 0.275 & LOW \\
\midrule
\multicolumn{7}{l}{\emph{Anti-symmetry ablation (best)}} \\
$\tau{=}0.65$, $\lambda{=}0.1$ & 0.483 & 0.555 & 0.124 & 0.224 & 0.288 & LOW \\
$\tau{=}0.5$, $\lambda{=}0.5$  & 0.484 & 0.528 & 0.127 & 0.241 & 0.313 & LOW \\
$\tau{=}0.3$, $\lambda{=}0.5$  & 0.498 & 0.560 & 0.139 & 0.249 & 0.281 & LOW \\
\midrule
\multicolumn{7}{l}{\emph{ArcFace fine-tuned}} \\
ArcFace+MiewID-MSv2    & 0.458 & 0.472 & 0.148 & 0.314 & 0.348 & LOW \\
ArcFace+DINOv3-ViT     & 0.458 & 0.448 & 0.244 & 0.545 & 0.792 & LOW \\
\bottomrule
\end{tabular}}
\end{table}

\subsection{Comprehensive Inpainted BG/FG Evaluation (102 Models)}\label{sec:supp_bgfg_102}

We also ran the inpainted \BGFG{} diagnostic across a 102-model suite covering
frozen baselines, ArcFace and triplet benchmarks, anti-symmetry ablations,
Lorentz finalists, and a small set of additional baselines.  The inpainted
\BGFG{} ratio has mean 0.396 (median 0.324), ranges from 0.120 to 1.166, and is
LOW ($<0.95$) for 98/102 models, MEDIUM (0.95--1.10) for 2/102, and HIGH
($>1.10$) for 2/102.  Only three models exceed $\BGFG \ge 1.0$:
EfficientNetV2-M (1.034), meFeM-B (1.125), and I-JEPA (1.166).

\begin{table}[t]
\centering
\caption{Summary statistics across all 102 evaluated models.
Legacy \BGSFG{} uses the na\"ive
\texttt{background\_only} construction; inpainted \BGFG{} uses the
leakage-controlled \texttt{background\_inpainted} variant.}
\label{tab:supp_bgfg_102_stats}
\small
\begin{tabular}{@{}lrrrrrr@{}}
\toprule
Metric & Mean & Median & Min & Max & \#${\ge}1$ & \#${>}1.1$ \\
\midrule
Legacy \BGSFG{} & 0.501 & 0.440 & 0.172 & 1.355 & 12 & 7 \\
Inpainted \BGFG{} & 0.396 & 0.324 & 0.120 & 1.166 & 3 & 2 \\
Silhouette/FG & 0.366 & 0.348 & 0.125 & 0.860 & 0 & 0 \\
\bottomrule
\end{tabular}
\end{table}

All 102 models have Silhouette/FG $< 0.95$ (max 0.860), so silhouette/shape
alone never matches coat-pattern performance.

Across the same 102 models, \BGSFG{} and inpainted \BGFG{} rank models
almost identically (Spearman $\rho{=}0.964$).  The main difference is scale:
the zeroed-foreground construction inflates ratios for frozen baselines because the
silhouette-shaped hole is an easy cue.  Inpainted \BGFG{} is also strongly
correlated with Silhouette/FG (Spearman $\rho{=}0.914$).  Full/FG is only
moderately correlated with inpainted \BGFG{} (Spearman $\rho{=}0.490$), since
Full/FG asks whether the standard input helps, not whether background alone
carries identity signal.

For the triplet benchmark, we denote the baseline triplet objective as T0 and
the anti-symmetry-regularised variant as T1
(see Sec.~\ref{sec:supp_triplet_bench} for full triplet results and definitions).
T1 shifts inpainted \BGFG{}
upward relative to T0: 0.298 (T0 mean) versus 0.407 (T1 mean), a +0.109 change.
Both remain well below the low-risk threshold, but the shift is a reminder that
laterality-targeted regularisation can move the context axis too.  That is why
we keep reporting both diagnostics.

%\paragraph{Auto-generated sensitivity tables.}
%\IfFileExists{compute_reproducibility/bgfg_sensitivity_tables.tex}{%
%\input{compute_reproducibility/bgfg_sensitivity_tables.tex}
%}{%
%\noindent\textit{Sensitivity tables are generated by
%\texttt{analyze\_bgfg\_sensitivity.py} and inserted at build time when
%\texttt{compute\_reproducibility/bgfg\_sensitivity\_tables.tex} is present.}
%}

% =====================================================================
\section{Three-Tier Negative Mining Strategy}\label{sec:supp_mining}

\begin{table}[t]
\centering
\caption{Three-tier negative mining strategy linking diagnostics to training.}
\label{tab:supp_mining}
\begin{tabular}{@{}clll@{}}
\toprule
Tier & Negative Type & Defeats & Origin \\
\midrule
1 (Hard)      & Mirror (LR flip)              & Laterality shortcut    & Axis~2 \\
1 (Hard)      & Background cutout             & Non-coat context memorisation & Axis~1 \\
2 (Semi-hard) & Closest different individual  & Confusable pairs       & Offline ranking \\
\bottomrule
\end{tabular}
\end{table}

\Cref{tab:supp_mining} details the three-tier negative mining strategy
described in the main paper (Sec.~3.4).  Tier~1 negatives directly
target the two shortcut axes: mirror flips defeat laterality shortcuts
(Axis~2) and background cutouts target non-coat context memorisation
(Axis~1; \BGSFG{}).
Tier~2 semi-hard negatives use offline embedding ranking to identify
the closest different-identity pairs, targeting confusable individuals.

% =====================================================================
\section{Euclidean Mitigation Objectives}\label{sec:supp_euclidean_obj}

This section collects the Euclidean objectives referenced from the main paper.
These models are trained on foreground-only cutouts, and we use no horizontal
flip augmentation because mirrored jaguar images can corrupt identity
supervision under the laterality diagnostic.

\paragraph{Sub-Center ArcFace loss.}
Let $z = f(x) / \|f(x)\|$ be the $\ell_2$-normalised embedding and
$\{W_{c,j}\}_{j=1}^{k}$ the $k$ normalised sub-center weight vectors for
class~$c$.  For the ground-truth class~$y$, we select the closest sub-center
and apply an additive angular margin~$m$:
\begin{equation}\label{eq:arcface}
  \mathcal{L}_{\text{arcface}}
  = -\log
    \frac{e^{\,s\,\cos(\theta_{y}+m)}}
         {e^{\,s\,\cos(\theta_{y}+m)}
          + {\displaystyle\sum_{c \neq y}}\;
            e^{\,s\,\cos\theta_{c}}}\,,
  \quad
  \cos\theta_{c}
  = \max_{j \in \{1,\dots,k\}}
    z^{\!\top} W_{c,j}\,.
\end{equation}
We use $k{=}3$ sub-centers, angular margin $m{=}0.5$, and scale $s{=}51.5$.
Multiple sub-centers let the loss absorb pose and occlusion variation without
forcing every image of an individual onto a single prototype~\cite{deng2020sub}.

\paragraph{Anti-symmetry regulariser.}
To reduce laterality shortcuts, we penalise high similarity between an image
and its horizontal flip:
\begin{equation}\label{eq:anti}
  \mathcal{L}_{\text{anti}}
  = \frac{1}{B}\sum_{i=1}^{B}
    \max\!\bigl(0,\;
      \cosim\bigl(f(I_i),\, f(\text{flip}(I_i))\bigr) - \tau
    \bigr)\,,
\end{equation}
with margin $\tau$.  Rather than pulling $(I, \text{flip}(I))$ together as a
standard augmentation-invariant pipeline would, we treat the pair as a soft
negative and push it apart. This is the symmetry we explicitly want to resist,
since networks can naturally learn transformed copies of features under
flips~\cite{olah2020naturally}.

\paragraph{Combined Euclidean objective.}
When anti-symmetry regularisation is used alongside ArcFace, the training
objective becomes
\begin{equation}\label{eq:total}
  \mathcal{L}_{\text{total}}
  = \mathcal{L}_{\text{arcface}}(z, y)
  + \lambda\,\mathcal{L}_{\text{anti}}(f(I), f(\text{flip}(I)))\,.
\end{equation}
The ablation in \cref{sec:supp_antisym_ablation} sweeps
$\tau$ and $\lambda$ on the fixed validation coreset.

% =====================================================================
\section{Lorentz Hyperbolic Benchmark}\label{sec:supp_lorentz}

\begin{table}[t]
\centering
\caption{Lorentz hyperbolic benchmark (validation mAP, mean $\pm$ std
across 3 seeds).  O1 adds mirror-negative loss.  Bold: best per backbone.}
\label{tab:supp_lorentz}
\begin{tabular}{@{}lccr@{}}
\toprule
Backbone & O0 (standard) & O1 (mirror neg) & $\Delta$ \\
\midrule
MiewID-MSv3        & $0.537 \pm 0.034$ & $\mathbf{0.608 \pm 0.007}$ & $+$13.3\% \\
MiewID-MSv2        & $0.532 \pm 0.002$ & $\mathbf{0.565 \pm 0.024}$ & $+$6.0\% \\
ConvNeXt-tiny      & $0.326 \pm 0.003$ & $\mathbf{0.328 \pm 0.008}$ & $+$0.6\% \\
Swin-tiny          & $\mathbf{0.312 \pm 0.006}$ & $0.310 \pm 0.005$ & $-$0.7\% \\
EfficientNetV2     & $\mathbf{0.305 \pm 0.005}$ & $0.301 \pm 0.006$ & $-$1.2\% \\
MegaDescriptor-L   & $\mathbf{0.243 \pm 0.008}$ & $0.239 \pm 0.005$ & $-$1.6\% \\
\bottomrule
\end{tabular}
\end{table}

\Cref{tab:supp_lorentz} reports the full Lorentz hyperbolic benchmark
across 6~backbones $\times$ 2~objectives $\times$ 3~seeds.
Mirror-negative training (O1) selectively improves wildlife-pretrained
backbones: MiewID-MSv3 gains 13.3\% and MiewID-MSv2 gains 6.0\%, while
generic backbones (ConvNeXt, Swin, EfficientNetV2, MegaDescriptor)
show $<$1\% change; laterality-aware features from
wildlife-specific pre-training appear necessary for the
mirror-negative signal to take effect.

\subsection{Lorentz Loss Formulation}\label{sec:supp_lorentz_loss}

We define the three components of the Lorentz training objective.

\paragraph{Solidity-to-variance mapping.}
Mask solidity $s \in [0,1]$ is computed from the SAM~3 mask as the
ratio of mask area to convex-hull area (see \cref{sec:supp_masks,eq:solidity_def}).
We map it to per-sample variance:
\begin{equation}\label{eq:sigma2}
  \sigma^{2}(s) = \sigma_{\min}^{2}
    + \bigl(\sigma_{\max}^{2} - \sigma_{\min}^{2}\bigr)\,(1-s)^{\gamma}\,,
\end{equation}
with $\sigma_{\min}{=}0.25$, $\sigma_{\max}{=}0.9$, $\gamma{=}1.5$.
Low-solidity (fragmented) masks receive high variance, encoding epistemic
uncertainty about identity.

\paragraph{Probabilistic supervised contrastive loss~\cite{khosla2020supervised}.}
Let $\mu_i$ denote the Lorentz embedding of sample~$i$ and $d(\cdot,\cdot)$
the geodesic distance on the Lorentz manifold.  We define a probabilistic
score:
\begin{equation}\label{eq:supcon_score}
  \ell_{ij} = \frac{-\,d^{2}(\mu_i, \mu_j)}
               {2\,(\sigma^{2}_i + \sigma^{2}_j)\,\tau}\,,
\end{equation}
where $\tau{=}0.1$ is a temperature.  The contrastive loss averages over all
anchors with at least one same-identity partner:
\begin{equation}\label{eq:supcon}
  \mathcal{L}_{\text{supcon}}
  = -\frac{1}{|A|}\sum_{i \in A}\;
    \frac{1}{|P_i|}\sum_{j \in P_i}
      \!\left[\,\ell_{ij}
        - \log\!\sum_{n \neq i} e^{\,\ell_{in}}\right],
\end{equation}
where $A$ is the set of valid anchors and $P_i = \{j : y_j{=}y_i,\, j \neq i\}$.

\begin{figure}[t]
\centering
\begin{minipage}[t]{0.48\textwidth}
  \centering
  \includegraphics[width=\textwidth]{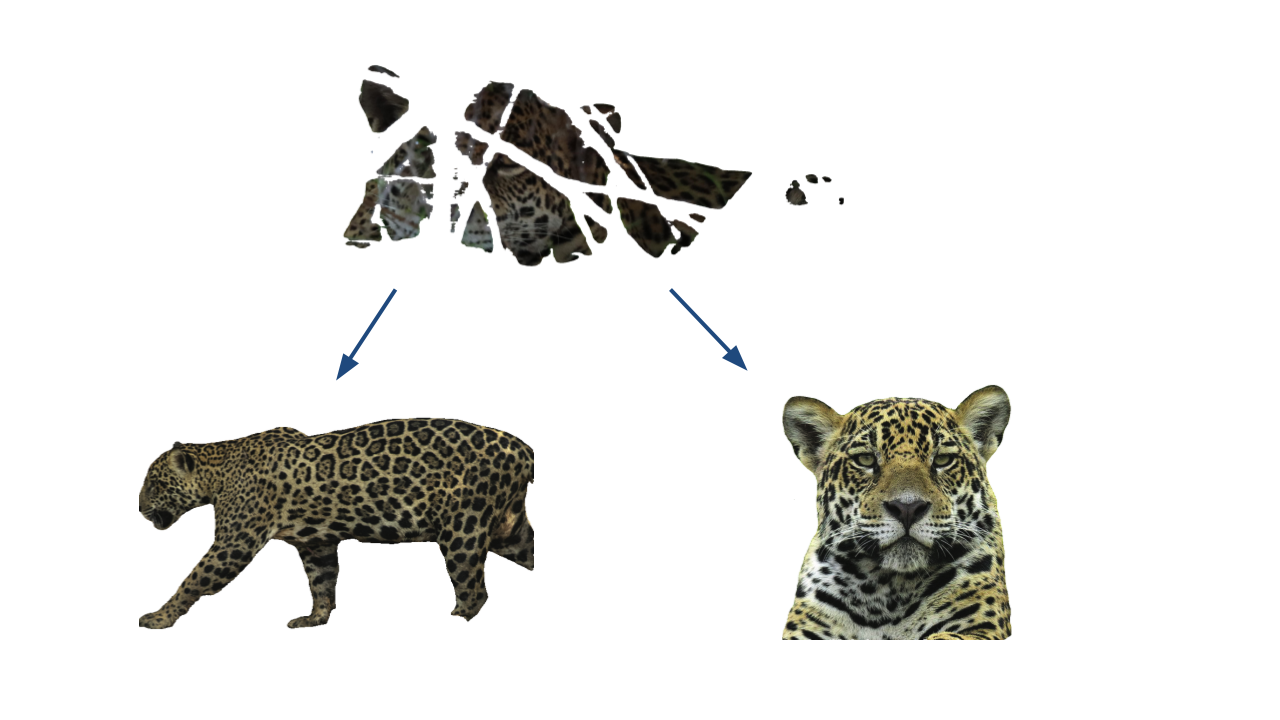}
  \caption{\textbf{Solidity-based quality hierarchy.}  A low-solidity
  (fragmented) mask at top yields a parent embedding with higher variance.
  High-quality full-body and close-up crops (bottom) produce tighter child
  embeddings on the Lorentz manifold.}
  \label{fig:supp_solidity}
\end{minipage}\hfill
\begin{minipage}[t]{0.48\textwidth}
  \centering
  \includegraphics[width=\textwidth]{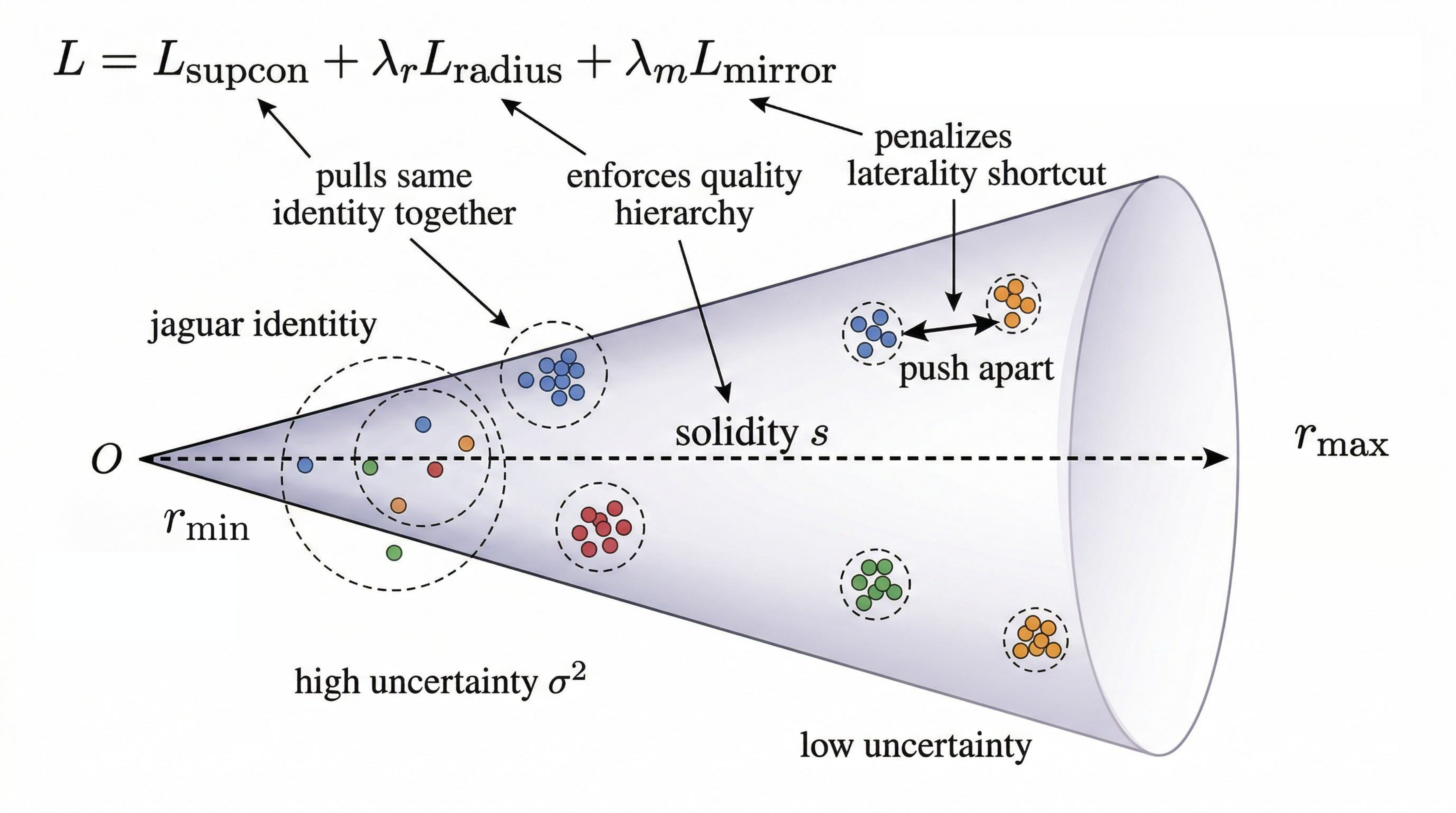}
  \caption{\textbf{Lorentz cone entailment geometry.}  Geodesic radius
  $r=d(\mu,\mathbf{o})$ encodes specificity: low-solidity images (high
  variance $\sigma^2(s)$; Eq.~\ref{eq:sigma2}) lie nearer the origin and
  define broad uncertainty cones, while high-solidity images are pushed
  toward larger target radii $g(s)$ (Eq.~\ref{eq:target_radius}).  The
  radius prior (Eq.~\ref{eq:radius}) implements this hierarchy by penalising
  deviations from $g(s)$ under solidity-weighted uncertainty.}
  \label{fig:supp_lorentz_cone}
\end{minipage}
\end{figure}

\paragraph{Radius prior loss (entailment).}
The radius prior encourages each embedding's distance from the Lorentz
origin to match a solidity-derived target, enforcing the cone entailment
hierarchy:
\begin{equation}\label{eq:radius}
  \mathcal{L}_{\text{radius}}
  = \frac{1}{N}\sum_{i=1}^{N}
    \left[\frac{(r_i - g(s_i))^{2}}{2\,\kappa\,\sigma^{2}_i}
      + \tfrac{1}{2}\log(\kappa\,\sigma^{2}_i)\right],
\end{equation}
where $r_i = d(\mu_i, \mathbf{o})$ is the geodesic radius from the origin,
and the target radius is:
\begin{equation}\label{eq:target_radius}
  g(s) = r_{\min} + (r_{\max} - r_{\min})\cdot s\,,
\end{equation}
with $r_{\min}{=}0.2$, $r_{\max}{=}2.0$, $\kappa{=}2.0$.  High-solidity
images are pushed further from the origin (tight identity clusters);
low-solidity images remain near the origin with high uncertainty, forming
parent nodes in the entailment hierarchy (\cref{fig:supp_lorentz_cone}).

\paragraph{Mirror-negative loss.}
To penalise laterality collapse in hyperbolic space:
\begin{equation}\label{eq:mirror_hyp}
  \mathcal{L}_{\text{mirror}}
  = \frac{1}{N}\sum_{i=1}^{N}
    \max\!\bigl(0,\; m_{\text{mirror}} - d(\mu_i, \mu_i^{\text{flip}})\bigr)\,,
\end{equation}
with $m_{\text{mirror}}{=}0.5$.

\paragraph{Combined objective.}
\begin{equation}\label{eq:lorentz}
  \mathcal{L}_{\text{Lorentz}}
  = \mathcal{L}_{\text{supcon}}
  + \lambda_{r}\,\mathcal{L}_{\text{radius}}
  + \lambda_{m}\,\mathcal{L}_{\text{mirror}}\,,
\end{equation}
with $\lambda_{r}{=}0.15$ and $\lambda_{m}{=}0.1$ (O1) or
$\lambda_{m}{=}0$ (O0 baseline).

\subsection{Lorentz Training Details (Implementation)}

We summarise the core Lorentz training hyperparameters and selection criteria
used for the benchmark in Sec.~\ref{sec:supp_lorentz}.  Unless otherwise
stated, we train for up to 80 epochs with early stopping on the fixed
355-image validation coreset and select the checkpoint with the lowest
validation total loss.

\paragraph{Model and embedding dimension.}
We use a $D{=}256$-dimensional Lorentz embedding with a 1024-dimensional
projection hidden layer, dropout 0.1, and GeM pooling ($p{=}3.0$ clamped to
$[1,6]$).  Curvature is fixed to 1.0 (not learned) and embeddings are projected
to the manifold with tangent-space clipping (norm 3.0).

\paragraph{Batch construction.}
Batches are identity-uniform: batch size 64 with 16 identities per batch
(4 images per identity) using an \texttt{IdentityUniformSampler} with
\texttt{drop\_last=True}.

\paragraph{Optimizer and learning-rate schedule.}
We optimise on the Lorentz manifold using Riemannian Adam
(\texttt{geoopt.optim.RiemannianAdam}).  Trainable parameters are split into
backbone and head groups with separate learning rates ($\text{lr}_{\text{head}}{=}3{\times}10^{-4}$,
$\text{lr}_{\text{backbone}}{=}10^{-4}$ when unfrozen) and weight decay $10^{-4}$
applied to non-bias/non-normalisation parameters.  The learning rate uses a
5-epoch linear warmup followed by cosine decay to $\text{lr}_{\min}{=}10^{-6}$.
Gradients are clipped to norm 5.0.

\paragraph{Augmentations and inputs.}
Images are resized to the backbone's canonical input size and normalised with
ImageNet statistics.  The default augmentation profile applies
RandomAffine ($\pm 15^\circ$, translate 0.15, scale 0.9--1.1) and ColorJitter;
no horizontal flips are used.  Training can be run on \texttt{full\_rgb} or
\texttt{foreground\_only} cutouts (background zeroed by the alpha mask).

\paragraph{Checkpoint selection and finalists.}
We select the best checkpoint by lowest validation total loss on the held-out
validation coreset, and we report the best observed validation identity-balanced
mAP for reference.  For the 36-configuration benchmark suite
(6 backbones $\times$ 2 objectives $\times$ 3 seeds), finalists are selected by
mean validation identity-balanced mAP across seeds and then evaluated on the
held-out test set.

% =====================================================================
\section{ArcFace Backbone Benchmark}\label{sec:supp_arcface_bench}

\paragraph{Training details.}
ArcFace fine-tuning uses AdamW with linear warmup from $1.5{\times}10^{-5}$ to
$1.5{\times}10^{-3}$ for 15 epochs followed by exponential decay (rate 0.8 per
epoch), weight decay $10^{-4}$, and batch size 64.  Images are resized to
$440{\times}440$ and augmented with CLAHE (clip limit 2.0; tile grid
$8{\times}8$; $p{=}0.5$), random sharpening ($\alpha \in [0.2, 0.5]$; $p{=}0.5$),
ShiftScaleRotate (shift 0.25; scale 0.20; rotation 15$^\circ$; $p{=}0.7$),
ColorJitter ($p{=}0.5$), and random grayscale ($p{=}0.1$), followed by ImageNet
normalisation.  We use no horizontal flip augmentation.  Validation uses the
fixed held-out coreset and we select the checkpoint with the lowest validation
loss (max 50 epochs; early-stopping patience 10).

\begin{table}[t]
\centering
\caption{ArcFace backbone benchmark (test macro mAP, mean $\pm$ std
across 3 seeds).  O0: ArcFace only; O1: ArcFace + anti-symmetry
($\tau{=}0.5$, $\lambda{=}0.1$).  All four backbones improve over
their frozen baselines (gains range from $+$38.7\% to $+$134.2\%).}
\label{tab:supp_arcface_bench}
\begin{tabular}{@{}lccccr@{}}
\toprule
Backbone & Frozen mAP & O0 (ArcFace) & O1 (ArcFace+AntiSym) & \BGSFG{} (O0) \\
\midrule
Swin-T          & 0.312 & $0.469 \pm 0.008$ & $0.475 \pm 0.008$ & $0.455 \pm 0.008$ \\
ConvNeXt-T      & 0.326 & $0.452 \pm 0.003$ & $0.450 \pm 0.016$ & $0.443 \pm 0.014$ \\
EffNetV2-M      & 0.193 & $0.452 \pm 0.014$ & $0.450 \pm 0.015$ & $0.341 \pm 0.005$ \\
MegaDesc-L      & 0.245 & $0.378 \pm 0.009$ & $0.363 \pm 0.016$ & $0.433 \pm 0.015$ \\
\bottomrule
\end{tabular}
\end{table}

\Cref{tab:supp_arcface_bench} reports ArcFace fine-tuning results
across four additional backbones (3 seeds each), alongside the
MiewID and DINOv3 results in the main paper.  Swin-T gains $+$50.3\%
over frozen, ConvNeXt-T $+$38.7\%, EfficientNetV2-M $+$134.2\%, and
MegaDescriptor-L $+$54.3\%.  Anti-symmetry regularisation (O1) has
little effect on these non-wildlife backbones ($\Delta < 1$~pp for
Swin and ConvNeXt, $-$1.5~pp for MegaDescriptor), matching the
Lorentz benchmark (\cref{tab:supp_lorentz}) where the mirror-negative
signal helps only wildlife-pretrained models.  \BGSFG{} ratios are low
across all configurations (0.34--0.46), so ArcFace training reduces
non-coat context reliance (lower \BGSFG{}) across backbones under the
zeroed-foreground diagnostic.

% Removed: head-to-head comparison table (redundant with main paper)
% \begin{table}[t]
% \centering
% \caption{Head-to-head comparison of the two main-paper ArcFace backbones.
% DINOv3 gives stronger ranking metrics, while MiewID remains more robust to
% background context.}
% \label{tab:head2head}
% \begin{tabular}{@{}lccc@{}}
% \toprule
% Metric & ArcFace+MiewID & ArcFace+DINOv3 & Winner \\
% \midrule
% mAP (macro)         & 0.458         & 0.458          & Tie \\
% CMC@1 (macro)       & 72.9\%        & \textbf{78.5\%} & DINOv3 ($+$7.7\%) \\
% CMC@5 (macro)       & 80.0\%        & \textbf{87.6\%} & DINOv3 ($+$9.5\%) \\
% \BGFG{}             & \textbf{0.314} & 0.545          & MiewID ($-$42.4\%) \\
% Tail CMC@1          & 61.1\%        & \textbf{70.4\%} & DINOv3 ($+$15.2\%) \\
% \bottomrule
% \end{tabular}
% \end{table}

% =====================================================================
\section{Anti-Symmetry Ablation}\label{sec:supp_antisym_ablation}

We sweep the anti-symmetry margin $\tau$ over $\{0.3,0.5,0.65\}$ and the loss
weight $\lambda$ over $\{0.1,0.3,0.5\}$ for ArcFace+MiewID-MSv2 on the fixed
validation coreset.  We evaluate all nine settings on the held-out test set.
The table below gives the full grid summarized briefly in the main paper.

\begin{table}[t]
\centering
\caption{Anti-symmetry ablation ($3{\times}3$ grid).  Baseline is
ArcFace+MiewID without $\mathcal{L}_{\text{anti}}$.  Hyperparameters are
selected on the validation coreset; test-set metrics are reported for
transparency.}
\label{tab:supp_antisym_ablation}
\begin{tabular}{@{}ccccc@{}}
\toprule
$\tau$ & $\lambda$ & mAP (macro) & CMC@1 (macro) & \BGFG{} \\
\midrule
\multicolumn{2}{c}{\emph{baseline}} & 0.458 & 72.9\% & 0.314 \\
\midrule
0.3  & 0.1 & 0.497 & 75.1\% & 0.255 \\
0.3  & 0.3 & 0.480 & 74.0\% & 0.244 \\
0.3  & 0.5 & \textbf{0.498} & 74.9\% & 0.249 \\
0.5  & 0.1 & 0.462 & 70.5\% & 0.249 \\
0.5  & 0.3 & 0.480 & 72.2\% & 0.244 \\
0.5  & 0.5 & 0.484 & 74.4\% & 0.241 \\
0.65 & 0.1 & 0.483 & 74.1\% & \textbf{0.224} \\
0.65 & 0.3 & 0.489 & \textbf{76.1\%} & 0.245 \\
0.65 & 0.5 & 0.447 & 72.6\% & 0.258 \\
\bottomrule
\end{tabular}
\end{table}

\Cref{tab:supp_antisym_ablation} reports the full sweep.  Eight of nine
settings improve over the ArcFace+MiewID baseline on macro mAP.  The selected
setting ($\tau{=}0.3$, $\lambda{=}0.5$) gives the best test macro mAP at 0.498,
while $\tau{=}0.65$, $\lambda{=}0.1$ yields the lowest \BGFG{} at 0.224.
Across the grid, \BGFG{} stays in a narrow range (0.224--0.258), so the
regulariser does not increase background reliance.

% =====================================================================
\section{Triplet Loss Benchmark}\label{sec:supp_triplet_bench}

\paragraph{Training details.}
Triplet fine-tuning uses TripletMarginLoss (L2 distance, margin 0.4) with
offline hard-negative triplets loaded from \texttt{triplets.json}, the same
6~backbones and 3~seeds as the ArcFace (\cref{sec:supp_arcface_bench}) and
Lorentz (\cref{sec:supp_lorentz}) benchmarks.  The baseline objective is
denoted \textbf{T0} (triplet loss only); the regularised variant \textbf{T1}
adds the same \texttt{AntiSymmetryLoss} used in the ArcFace O1 configuration
($\lambda{=}0.1$, $\tau{=}0.5$), applied to horizontally-flipped anchors.
Each mini-batch loads anchor, positive, and negative images (batch size 21;
each sample loads 3~images), so T0 sees $3\times$ as many images per step as
standard classification.  Unlike ArcFace and Lorentz, checkpoints are selected
by highest validation identity-balanced mAP.

\begin{table}[t]
\centering
\caption{Triplet loss benchmark (test macro mAP, mean $\pm$ std across 3 seeds).
T0: triplet only; T1: triplet + anti-symmetry ($\tau{=}0.5$, $\lambda{=}0.1$).
FG~mAP is foreground-only; \BGFG{} is the inpainted leakage-controlled ratio.}
\label{tab:supp_triplet_bench}
\resizebox{\textwidth}{!}{%
\begin{tabular}{@{}lcccccc@{}}
\toprule
& \multicolumn{3}{c}{T0 (Triplet only)} & \multicolumn{3}{c}{T1 (Triplet + AntiSym)} \\
\cmidrule(lr){2-4}\cmidrule(lr){5-7}
Backbone & Full mAP & FG mAP & \BGFG{} & Full mAP & FG mAP & \BGFG{} \\
\midrule
MiewID-MSv3   & $0.134 \pm 0.014$ & $0.549 \pm 0.006$ & $0.148 \pm 0.012$ & $\mathbf{0.510 \pm 0.003}$ & $0.574 \pm 0.005$ & $0.312 \pm 0.009$ \\
MiewID-MSv2   & $0.088 \pm 0.009$ & $0.531 \pm 0.007$ & $0.136 \pm 0.019$ & $0.458 \pm 0.008$ & $0.554 \pm 0.012$ & $0.280 \pm 0.005$ \\
EffNetV2-M    & $0.407 \pm 0.003$ & $0.495 \pm 0.004$ & $0.456 \pm 0.008$ & $0.348 \pm 0.005$ & $0.436 \pm 0.001$ & $0.511 \pm 0.009$ \\
Swin-T        & $0.379 \pm 0.008$ & $0.493 \pm 0.005$ & $0.472 \pm 0.004$ & $0.338 \pm 0.012$ & $0.503 \pm 0.006$ & $0.485 \pm 0.007$ \\
ConvNeXt-T    & $0.337 \pm 0.007$ & $0.502 \pm 0.003$ & $0.372 \pm 0.003$ & $0.322 \pm 0.006$ & $0.524 \pm 0.007$ & $0.370 \pm 0.002$ \\
MegaDesc-L    & $0.275 \pm 0.006$ & $0.596 \pm 0.001$ & $0.204 \pm 0.095$ & $0.249 \pm 0.004$ & $0.464 \pm 0.003$ & $0.483 \pm 0.003$ \\
\bottomrule
\end{tabular}}
\end{table}

\paragraph{Results.}
\Cref{tab:supp_triplet_bench} reports the full triplet benchmark.
For the active v5 claim bundle we keep both MSv3 triplet objectives in play
rather than forcing one single triplet row.  T0 is stronger on full-RGB
macro mAP (0.510 vs.\ 0.479) and cross-flank macro mAP (0.681 vs.\ 0.647),
while T1 is much better on Axis~2 under the rank-primary calibration view and
on raw mirror similarity.  The paired AP tests only support T0 over T1 on the
two inpainted vegetation endpoints, plus the laterality-axis differences in raw
mirror similarity and danger margin.  Cross-flank AP and all paired CMC
endpoints remain unsupported.  That split is the reason both rows stay in the
paper.

\begin{table}[t]
\centering
\caption{Cross-flank retrieval summary for the active v5 family-level bundle
(foreground-only, macro metrics).  Queries without a positive in the relevant
gallery are excluded from that split.  We report full cross-flank mAP and
CMC@1/5/10 together with the cross/within ratio.}
\label{tab:supp_crossflank_finalists}
\small
\resizebox{\textwidth}{!}{%
\begin{tabular}{@{}lccccc@{}}
\toprule
Model & X-mAP & X-CMC@1 & X-CMC@5 & X-CMC@10 & X/W \\
\midrule
ArcFace O0 (MSv3) & 0.576 $\pm$ 0.049 & 60.7\% $\pm$ 4.9\% & 69.0\% $\pm$ 4.3\% & 75.1\% $\pm$ 2.6\% & 0.777 $\pm$ 0.064 \\
ArcFace O1 (MSv3) & 0.566 $\pm$ 0.026 & 62.4\% $\pm$ 1.8\% & 69.9\% $\pm$ 4.7\% & 76.2\% $\pm$ 5.6\% & 0.780 $\pm$ 0.044 \\
Lorentz O0 (MSv3) & 0.689 $\pm$ 0.032 & 72.0\% $\pm$ 4.7\% & 81.5\% $\pm$ 0.7\% & 84.8\% $\pm$ 2.5\% & 0.899 $\pm$ 0.026 \\
Lorentz O1 (MSv2) & 0.653 $\pm$ 0.011 & 71.2\% $\pm$ 4.9\% & \textbf{84.4\% $\pm$ 1.2\%} & \textbf{88.8\% $\pm$ 1.0\%} & 0.881 $\pm$ 0.028 \\
Lorentz O1 (MSv3) & \textbf{0.692 $\pm$ 0.010} & 72.5\% $\pm$ 2.2\% & 81.9\% $\pm$ 3.4\% & 87.0\% $\pm$ 2.9\% & 0.876 $\pm$ 0.021 \\
Triplet T0 (MSv3) & 0.681 $\pm$ 0.013 & \textbf{73.0\% $\pm$ 2.3\%} & 81.3\% $\pm$ 2.4\% & 84.1\% $\pm$ 1.2\% & \textbf{0.923 $\pm$ 0.012} \\
Triplet T1 (MSv3) & 0.647 $\pm$ 0.003 & 71.5\% $\pm$ 0.4\% & 78.6\% $\pm$ 0.7\% & 83.2\% $\pm$ 0.7\% & 0.875 $\pm$ 0.004 
\\
\bottomrule
\end{tabular}}
\end{table}

% =====================================================================
\section{Paired Wilcoxon Summaries}\label{sec:supp_paired}

\begin{table}[t]
\centering
\small
\caption{Paired AP and Axis~2 summary from the active v5 bundle.  Rows use the
matched-objective comparison policy: same family, same variant, different
objective, complete seeds 42/43/44, two-sided Wilcoxon per seed, Fisher
combination, and Holm correction.}
\label{tab:supp_paired_ap}
\begin{tabular}{@{}p{0.23\textwidth}p{0.27\textwidth}p{0.40\textwidth}@{}}
\toprule
Comparison & Supported endpoints & Remaining tested endpoints \\
\midrule
ArcFace O0 vs O1 (MSv3) & axis2 danger margin & All Axis-1 AP endpoints, raw mirror similarity, and cross-flank AP are not supported after Fisher combination and Holm correction. \\
Lorentz O0 vs O1 (MSv3) & none & Full-RGB AP, foreground AP, silhouette AP, inpainted AP, raw mirror similarity, danger margin, and cross-flank AP are all not supported. \\
Triplet T0 vs T1 (MSv3) & inpainted vegetation 1 AP; inpainted vegetation 2 AP; axis2 danger margin; axis2 mirror similarity & Full-RGB AP, foreground AP, silhouette AP, silhouette+bg AP, vegetation 3 AP, and cross-flank AP are not supported. 
\\
\bottomrule
\end{tabular}
\end{table}

\begin{table}[t]
\centering
\small
\caption{Paired CMC summary from the same active v5 bundle.  No matched
objective pair has a supported CMC@1/5/10 difference after the full
Wilcoxon-Fisher-Holm pipeline.}
\label{tab:supp_paired_cmc}
\begin{tabular}{@{}p{0.23\textwidth}p{0.13\textwidth}p{0.54\textwidth}@{}}
\toprule
Comparison & Supported endpoints & Note \\
\midrule
ArcFace O0 vs O1 (MSv3) & none & All full-RGB, foreground, inpainted, silhouette, silhouette+bg, and cross-flank CMC@1/5/10 endpoints are not supported. \\
Lorentz O0 vs O1 (MSv3) & none & All full-RGB, foreground, inpainted, silhouette, silhouette+bg, and cross-flank CMC@1/5/10 endpoints are not supported. \\
Triplet T0 vs T1 (MSv3) & none & All full-RGB, foreground, inpainted, silhouette, silhouette+bg, and cross-flank CMC@1/5/10 endpoints are not supported. 
\\
\bottomrule
\end{tabular}
\end{table}

% =====================================================================
\section{Classical CV Baseline: HotSpotter}\label{sec:supp_hotspotter}

\begin{table}[t]
\centering
\caption{Classical CV baseline.  HotSpotter+VLAD shows low shortcut reliance
under both metrics, with zero learned parameters but laterality-blind.}
\label{tab:supp_hotspotter}
\begin{tabular}{@{}lcccccr@{}}
\toprule
Method & mAP & \BGFG{} & \BGSFG{} & Mirror Sim & Emb.\ Dim \\
\midrule
HotSpotter+VLAD & 0.288 & 0.535 & 0.845 & 1.000 & 8{,}192 \\
HotSpotter+Mean & 0.191 & 0.806 & 1.047 & 1.000 & 128 \\
\bottomrule
\end{tabular}
\end{table}

\Cref{tab:supp_hotspotter} reports the full HotSpotter evaluation.
VLAD aggregation~\cite{jegou2010vlad} with $K{=}64$ clusters yields an
8{,}192-dimensional descriptor with zero learned parameters.  VLAD
improves mAP by 51\% over mean pooling (0.288 vs.\ 0.191) and reduces
$\BGSFG$ from 1.047 to 0.845.  However, both variants are
laterality-blind (mirror similarity $= 1.000$).

% =====================================================================
\section{ArcFace+DINOv3-ViT-S/16 Detailed Results}\label{sec:supp_dino}

\begin{table}[t]
\centering
\caption{ArcFace+DINOv3-ViT-S/16 results.  Self-supervised features match
wildlife-specific features on mAP with superior CMC@1 and parameter
efficiency.}
\label{tab:supp_arcface_dino}
\begin{tabular}{@{}lccc@{}}
\toprule
Metric & Frozen DINOv3 & ArcFace+DINOv3 & Change \\
\midrule
Test mAP (macro)        & 0.298 & 0.458 & $+$53.7\% \\
Test mAP (micro)        & 0.270 & 0.479 & $+$77.4\% \\
CMC@1 (macro)           & 70.0\% & 78.5\% & $+$12.1\% \\
CMC@1 (micro)           & 72.8\% & 82.2\% & $+$12.9\% \\
CMC@5 (macro)           & 77.6\% & 87.6\% & $+$12.9\% \\
CMC@5 (micro)           & 82.2\% & 89.8\% & $+$9.2\% \\
CMC@10 (macro)          & 82.2\% & 90.0\% & $+$9.5\% \\
CMC@10 (micro)          & 86.8\% & 92.2\% & $+$6.2\% \\
\BGSFG{} Ratio           & 1.020 & 0.792 & $-$22.4\% \\
Foreground-only mAP     & 0.296 & 0.448 & $+$51.4\% \\
Background-only mAP     & 0.302 & 0.355 & $+$17.5\% \\
\bottomrule
\end{tabular}
\end{table}

\Cref{tab:supp_arcface_dino} provides the single-seed (seed~42) metric breakdown for
ArcFace+DINOv3-ViT-S/16.  This model achieves identical macro mAP
(0.458) to ArcFace+MiewID while using $11.5\times$ fewer trainable
parameters (826K vs.\ 9.48M).  Its \BGSFG{} ratio of 0.792 is a
22.4\% reduction from the frozen DINOv3 baseline (1.020); ArcFace
fine-tuning reduces non-coat context reliance (lower \BGSFG{}) across
backbone architectures.

% =====================================================================
\section{Mirror Similarity Ranking}\label{sec:supp_mirror}

\begin{table}[t]
\centering
\caption{Mirror similarity ranking (15 models).  Wildlife pre-training
(MiewID) is the dominant factor; self-supervised models exceed 0.96.
Standard deviations computed across 371 test images; T1 models show the
highest per-image variance (std $> 0.06$), indicating identity-dependent
laterality sensitivity.}
\label{tab:supp_mirror}
\begin{tabular}{@{}rlcccl@{}}
\toprule
Rank & Model & Mirror Sim & Std & \BGSFG{} & Tier \\
\midrule
1  & MiewID-MSv3 (frozen)       & \textbf{0.746} & 0.067 & NA    & T1: Laterality-Aware \\
2  & MiewID-MSv2 (frozen)       & \textbf{0.752} & 0.064 & 0.850 & T1: Laterality-Aware \\
3  & ArcFace+MiewID-MSv2        & \textbf{0.833} & 0.083 & 0.348 & T1: Laterality-Aware \\
4  & MegaDescriptor             & 0.914 & 0.035 & 1.020 & T2: Moderate \\
5  & ResNet50                   & 0.958 & 0.021 & 1.125 & T2: Moderate \\
6  & DINOv3-ViT CLS             & 0.963 & 0.012 & 1.020 & T2: Moderate \\
7  & I-JEPA                     & 0.968 & 0.011 & 1.27  & T3: Strong Shortcut \\
8  & DINOv3-ViT Pooled          & 0.979 & 0.008 & 1.179 & T3: Strong Shortcut \\
9  & DINOv3-ConvNeXt Pooled     & 0.981 & 0.015 & 1.355 & T3: Strong Shortcut \\
10 & EfficientNetV2-M           & 0.981 & 0.015 & 1.009 & T3: Strong Shortcut \\
11 & DINOv3-ConvNeXt CLS        & 0.984 & 0.015 & 1.161 & T3: Strong Shortcut \\
12 & ConvNeXtV2-Base            & 0.985 & 0.018 & 1.058 & T3: Strong Shortcut \\
13 & C-RADIO-v4                 & 0.997 & 0.002 & 1.22  & T4: Near-Perfect Sym. \\
14 & EVA-02                     & 0.997 & 0.004 & 0.930 & T4: Near-Perfect Sym. \\
15 & HotSpotter+VLAD             & 1.000 & 0.000 & 0.845 & T4: Near-Perfect Sym. \\
\bottomrule
\end{tabular}
\end{table}

\Cref{tab:supp_mirror} provides the full 15-model mirror similarity
ranking.  This table contains only non-Lorentz models, so Axis~2 is computed
with each model's native descriptor score; for the deep baselines that reduces
to cosine on the extracted embeddings.  The gap between MiewID ($0.752$) and
the next-best non-MiewID model (MegaDescriptor, $0.914$) is $0.162$, larger
than the gap between MegaDescriptor and the worst model, which points to
wildlife-specific pre-training as the dominant factor in laterality awareness.

% =====================================================================
\section{Auxiliary Mirrored-Query Retrieval Analysis}\label{sec:supp_correlation}

\begin{table}[t]
\centering
\caption{Rank correlation between inpainted \BGFG{} and the mirrored-query
retrieval ratio $r=\text{mAP}_{\text{mirror}}/\text{mAP}_{\text{regular}}$
(foreground-only cutouts) across the $N{=}15$ frozen models used in the main
paper.  This is an auxiliary retrieval stress test, not the canonical
definition of Axis~2.  Bootstrap 95\% CIs are percentile intervals with
$B{=}20{,}000$ resamples (seed~0).}
\label{tab:supp_independence}
\begin{tabular}{@{}lccc@{}}
\toprule
Correlation Metric & Value & $p$-value & 95\% CI \\
\midrule
Spearman $\rho$    & $-0.200$ & $0.493$ & $[-0.757,\,0.404]$ \\
Kendall $\tau$     & $-0.187$ & $0.388$ & $[-0.643,\,0.284]$ \\
\bottomrule
\end{tabular}
\end{table}

Table~\ref{tab:supp_independence} reports an auxiliary retrieval-level check
of leakage-controlled \BGFG{} against the mirrored-query ratio $r$.  Axis~2
itself remains mean mirror similarity under the model's native retrieval
score.  We report $r$ only as a downstream stress test, not as a second
definition of the axis.  We do not see a clear monotonic association across
the frozen suite, and with $N{=}15$ the uncertainty remains wide.  Across all
102 models with both diagnostics, the correlation is also weak (Spearman
$\rho{=}0.089$, $p{=}0.372$; Kendall $\tau{=}0.066$, $p{=}0.327$).  The
102-checkpoint suite spans 16 backbones.  It includes the ArcFace benchmark
(36), triplet benchmark (36), frozen baselines (16), anti-symmetry ablations
(10), Lorentz finalists (3), and ArcFace+DINOv3 (1).
Since many checkpoints share backbones and differ only in objective or random
seed, we treat this as a descriptive sanity check rather than a fully
independent sample.

% =====================================================================
\section{Long-Tail Performance Analysis}\label{sec:supp_longtail}

\begin{table}[t]
\centering
\caption{Long-tail performance breakdown for ArcFace+DINOv3-ViT-S/16.
Tail identities (9\% of data) still achieve 70.4\% CMC@1.}
\label{tab:supp_longtail}
\begin{tabular}{@{}lrrccc@{}}
\toprule
Tier & \# IDs & \% Data & CMC@1 & CMC@5 & CMC@10 \\
\midrule
HEAD (top 33\%)      & 10 & 65\% & 86.2\% & 93.1\% & 95.4\% \\
MEDIUM (middle 33\%) & 11 & 26\% & 80.0\% & 87.0\% & 89.0\% \\
TAIL (bottom 33\%)   & 10 &  9\% & 70.4\% & 81.5\% & 85.2\% \\
\bottomrule
\end{tabular}
\end{table}

\Cref{tab:supp_longtail} breaks down ArcFace+DINOv3 performance by
identity frequency tier.  The HEAD-to-TAIL gap in CMC@1 is
15.8~percentage points (86.2\% vs.\ 70.4\%): rare
identities are harder to recognise but still serviceable.

\begin{figure}[t]
\centering
\includegraphics[width=0.75\textwidth]{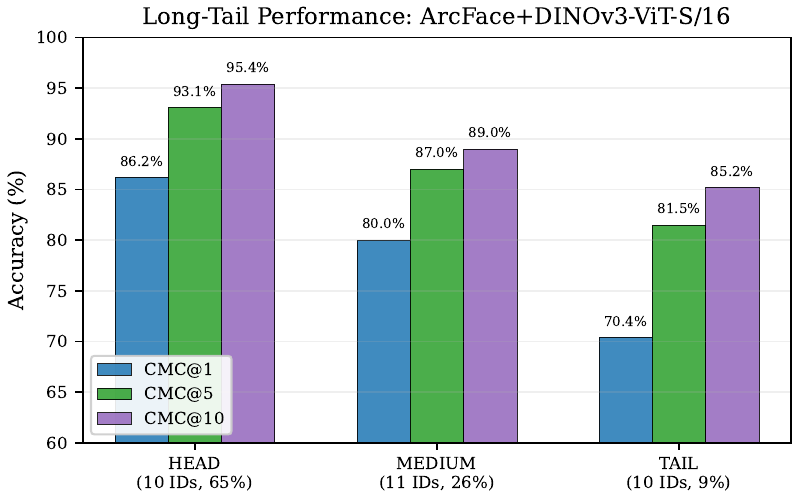}
\caption{\textbf{Long-tail performance breakdown for ArcFace+DINOv3.}
CMC@1/5/10 by identity frequency tier.  Tail identities (10 IDs, 9\% of
data) still achieve 70.4\% CMC@1, with a HEAD-to-TAIL gap of 15.8pp.}
\label{fig:supp_longtail_cmc}
\end{figure}

% =====================================================================
\section{Additional Visualisations}\label{sec:supp_figures}

\begin{figure}[t]
\centering
\includegraphics[width=0.8\textwidth]{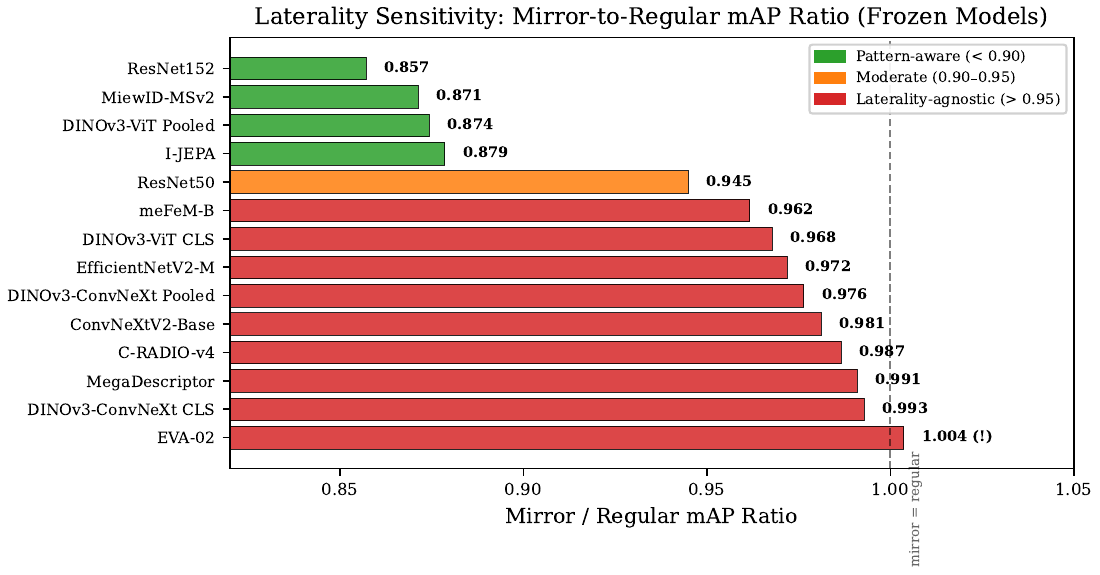}
\caption{\textbf{Auxiliary mirrored-query retrieval stress test.}
Mirror-to-regular mAP ratio for the 14 frozen models (foreground-only
cutouts).  This figure reports a retrieval-level check, not the canonical
definition of Axis~2, which remains mean mirror similarity under each model's
native retrieval score.  Lower values indicate greater asymmetry awareness.
Green: pattern-aware ($< 0.90$); orange: moderate ($0.90$--$0.95$); red:
laterality-agnostic ($> 0.95$).  The dashed line at $1.0$ marks the point
where mirrored queries perform as well as originals.  Only four models fall
below 0.90 (ResNet152, MiewID-MSv2, DINOv3-ViT Pooled, I-JEPA), while EVA-02
is mirror-preferred (ratio $> 1.0$).}
\label{fig:supp_laterality_ratio}
\end{figure}

\begin{table}[t]
\centering
\caption{Auxiliary mirrored-query retrieval ratio (mirror / regular queries)
on foreground-only images.  Numerical data behind
\cref{fig:supp_laterality_ratio}.  Axis~2 itself is mean mirror similarity
under the model's native retrieval score.  EVA-02 is the only model where
mirror queries retrieve better than originals (ratio $> 1.0$), consistent
with its near-perfect mirror similarity
($0.997$).}
\label{tab:supp_laterality_ratio}
\begin{tabular}{@{}rlcccl@{}}
\toprule
Rank & Model & Regular mAP & Mirror mAP & Ratio & Tier \\
\midrule
1  & ResNet152             & 0.215 & 0.184 & 0.857 & Pattern-Aware \\
2  & MiewID-MSv2            & 0.295 & 0.257 & 0.871 & Pattern-Aware \\
3  & DINOv3-ViT Pooled      & 0.278 & 0.243 & 0.874 & Pattern-Aware \\
4  & I-JEPA                 & 0.238 & 0.209 & 0.879 & Pattern-Aware \\
5  & ResNet50               & 0.227 & 0.215 & 0.945 & Moderate \\
6  & meFeM-B                & 0.199 & 0.192 & 0.962 & Lat.-Agnostic \\
7  & DINOv3-ViT CLS         & 0.295 & 0.286 & 0.968 & Lat.-Agnostic \\
8  & EfficientNetV2-M       & 0.198 & 0.193 & 0.972 & Lat.-Agnostic \\
9  & DINOv3-ConvNeXt Pooled & 0.211 & 0.206 & 0.976 & Lat.-Agnostic \\
10 & ConvNeXtV2-Base        & 0.209 & 0.205 & 0.981 & Lat.-Agnostic \\
11 & C-RADIO-v4             & 0.253 & 0.249 & 0.987 & Lat.-Agnostic \\
12 & MegaDescriptor         & 0.286 & 0.283 & 0.991 & Lat.-Agnostic \\
13 & DINOv3-ConvNeXt CLS    & 0.210 & 0.209 & 0.993 & Lat.-Agnostic \\
14 & EVA-02                 & 0.305 & 0.306 & \textbf{1.004} & Mirror-Preferred \\
\bottomrule
\end{tabular}
\end{table}

\begin{figure}[t]
\centering
\begin{subfigure}[t]{\textwidth}
  \centering
  \includegraphics[width=\textwidth]{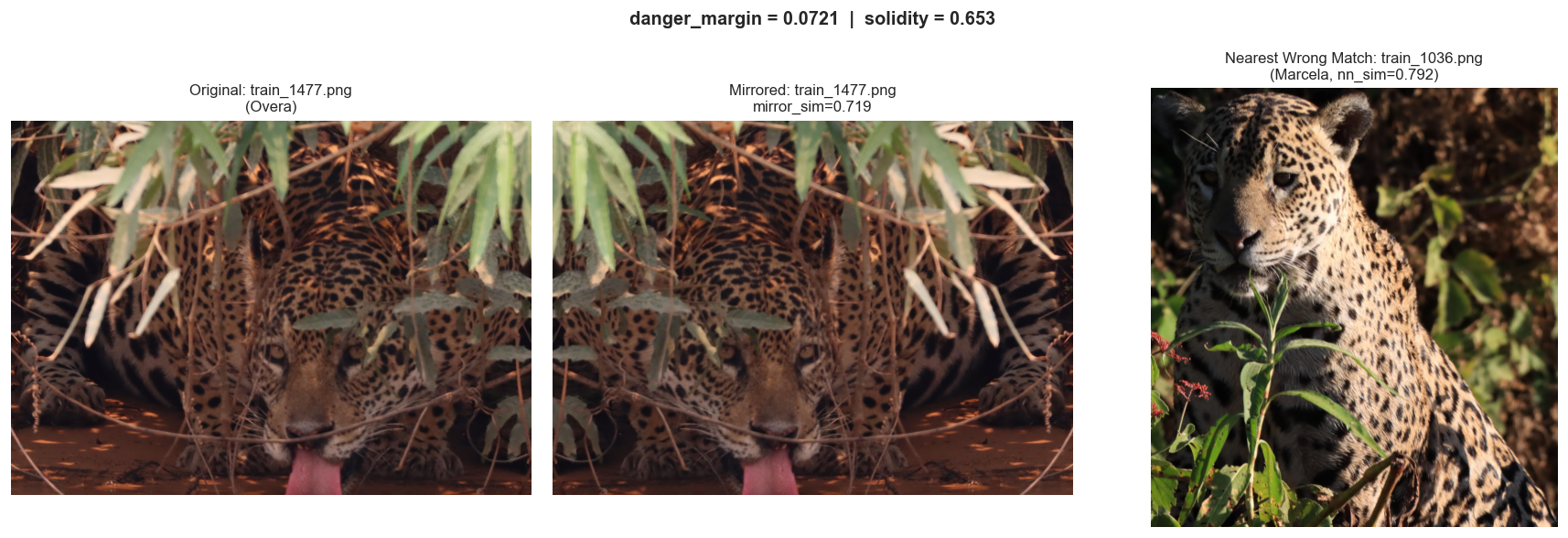}
  \caption{Overa (\texttt{train\_1477}): $\mirsim = 0.719$,
  $\nnsim = 0.792$ (Marcela), $\dangerm = +0.072$.}
  \label{fig:supp_asymmetry_overa}
\end{subfigure}
\vspace{0.3em}
\begin{subfigure}[t]{\textwidth}
  \centering
  \includegraphics[width=\textwidth]{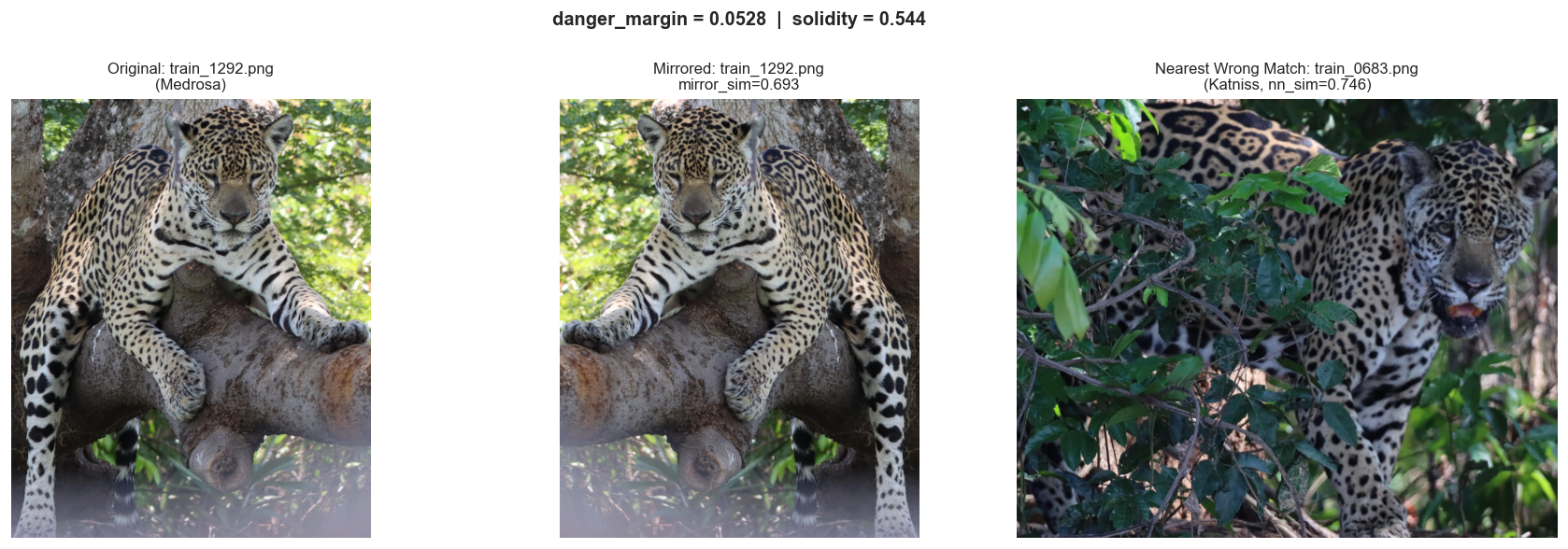}
  \caption{Medrosa (\texttt{train\_1292}): $\mirsim = 0.693$,
  $\nnsim = 0.746$ (Katniss), $\dangerm = +0.053$.}
  \label{fig:supp_asymmetry_medrosa}
\end{subfigure}
\caption{\textbf{Two positive-danger-margin cases according to MegaDescriptor-L embeddings.}
Each row: original foreground crop (left), horizontal mirror (centre),
nearest wrong-identity match (right).  In both cases the mirrored image is
\emph{less} similar to the original than a different individual
($\dangerm > 0$), confirming that horizontal flips can corrupt identity for
laterality-aware models.  Both images have below-average solidity
($0.65$ and $0.54$ vs.\ dataset mean $0.69$), suggesting poorer
segmentation increases vulnerability
(\cref{tab:supp_danger_margin}).}
\label{fig:supp_asymmetry_bias}
\end{figure}

\begin{table}[t]
\centering
\caption{Danger margin scan across all 1{,}895 training images
(MegaDescriptor).  Only 2~images (0.1\%) have positive danger margin,
both from low-contrast identities.  Identities sorted by max danger
margin; top~5 shown.  Aggregate: mean $\dangerm = -0.366$, median
$-0.369$.}
\label{tab:supp_danger_margin}
\begin{tabular}{@{}lrrccc@{}}
\toprule
Identity & Images & DM $> 0$ & Mean DM & Max DM \\
\midrule
Overa     & 62  & 1 (1.6\%) & $-$0.337 & $+$0.072 \\
Medrosa   & 170 & 1 (0.6\%) & $-$0.378 & $+$0.053 \\
Katniss   & 63  & 0         & $-$0.291 & $-$0.000 \\
Apeiara   & 20  & 0         & $-$0.370 & $-$0.013 \\
Ousado    & 179 & 0         & $-$0.356 & $-$0.034 \\
\midrule
\multicolumn{2}{l}{\emph{All 31 identities}} & 2 (0.1\%) & $-$0.366 & $+$0.072 \\
\bottomrule
\end{tabular}
\end{table}

\begin{table}[t]
\centering
\caption{Axis~2 summary from the active v5 TEST bundle.  Cross-space claims use
\texttt{mirror\_rank\_pct\_mean} (lower is better).  Raw mirror similarity and
z-normalised values are shown for context only and do not override the
rank-primary ordering.}
\label{tab:supp_axis2_mitigation}
\resizebox{\textwidth}{!}{%
\begin{tabular}{@{}lcccc@{}}
\toprule
Model & Raw mirror & Mirror rank pct & Mirror z & Mean $\dangerm$ \\
\midrule
ArcFace O0 (MSv3) & 0.855 $\pm$ 0.004 & 1.000 $\pm$ 0.000 & 0.798 $\pm$ 0.011 & -0.412 $\pm$ 0.006 \\
ArcFace O1 (MSv3) & 0.847 $\pm$ 0.002 & 0.667 $\pm$ 0.000 & 0.748 $\pm$ 0.002 & -0.391 $\pm$ 0.004 \\
Lorentz O0 (MSv3) & 0.290 $\pm$ 0.002 & 0.333 $\pm$ 0.471 & -0.260 $\pm$ 1.192 & -0.146 $\pm$ 0.003 \\
Lorentz O1 (MSv2) & 0.291 $\pm$ 0.002 & 0.333 $\pm$ 0.236 & -0.359 $\pm$ 0.615 & -0.118 $\pm$ 0.001 \\
Lorentz O1 (MSv3) & 0.296 $\pm$ 0.008 & 0.833 $\pm$ 0.236 & 0.619 $\pm$ 0.788 & -0.149 $\pm$ 0.008 \\
Triplet T0 (MSv3) & 0.741 $\pm$ 0.002 & 0.333 $\pm$ 0.000 & 0.124 $\pm$ 0.018 & -0.340 $\pm$ 0.004 \\
Triplet T1 (MSv3) & \textbf{0.435 $\pm$ 0.002} & \textbf{0.000 $\pm$ 0.000} & \textbf{-1.670 $\pm$ 0.005} & -0.108 $\pm$ 0.004 
\\
\bottomrule
\end{tabular}}
\end{table}

% =====================================================================
\section*{Implementation and Compute Details}\label{sec:supp_compute}

\paragraph{Frozen embedding extraction.}
\Cref{tab:supp_frozen_extraction} summarises the embedding extraction details
for the frozen-feature benchmark (main Table~1).  Unless otherwise noted, we
compute cosine similarity on $\ell_2$-normalised embeddings.

\begin{table}[t]
\centering
\caption{Frozen embedding extraction details for the 15 deep backbones from the
frozen benchmark (main Table~1; HotSpotter+VLAD excluded as classical CV).
Input sizes and global descriptor choices follow each model's
canonical evaluation wrapper.  ``HF processor'' denotes the HuggingFace image
processor; ``timm config'' denotes timm's resolved preprocessing configuration.}
\label{tab:supp_frozen_extraction}
\resizebox{\textwidth}{!}{%
\begin{tabular}{@{}lcccl@{}}
\toprule
Model & Input & Preprocess & Global descriptor & Dim \\
\midrule
DINOv3-ViT (CLS)           & 224 & HF processor & CLS token & 384 \\
DINOv3-ViT (pooled)        & 224 & HF processor & Mean pool patch tokens & 384 \\
DINOv3-ConvNeXt (CLS)      & 224 & HF processor & CLS token & 768 \\
DINOv3-ConvNeXt (pooled)   & 224 & HF processor & Mean pool patch tokens & 768 \\
MiewID-MSv2                & 440 & Resize + ImageNet norm & Model embedding (GeM + BN) & 2152 \\
MiewID-MSv3                & 440 & Resize + ImageNet norm & Model embedding (GeM + BN) & 2152 \\
MegaDescriptor-L           & 384 & Resize + ImageNet norm & Descriptor output & 1536 \\
EVA-02                     & 448 & timm config & Features (\texttt{num\_classes{=}0}) & 1024 \\
C-RADIO-v4                 & var. & Aspect-ratio preserve (max edge 768) & Summary tensor (L2 norm) & 2560 \\
I-JEPA                     & 224 & HF processor & Mean pool token features & 1280 \\
ResNet50                   & 224 & torchvision weights & Avgpool features & 2048 \\
ResNet152                  & 224 & torchvision weights & Avgpool features & 2048 \\
ConvNeXtV2-Base            & 224 & timm config & Features (\texttt{num\_classes{=}0}) & 1024 \\
EfficientNetV2-M           & 384 & torchvision weights & Avgpool features & 1280 \\
meFeM-B                    & 224 & timm config & CLS token & 768 \\
\bottomrule
\end{tabular}}
\end{table}

\paragraph{Compute.}
Experiments were run on (i) a local Apple Silicon machine (MPS) for
development and (ii) the HPI HPC cluster with NVIDIA H100 80GB HBM3 GPUs.
The frozen baseline benchmark comprises 14 backbones evaluated on 3 image
variants (\texttt{full\_rgb}, \texttt{foreground\_only},
\texttt{background\_inpainted}; 42 test-set evaluations).  The Lorentz benchmark trains
6 backbones $\times$ 2 objectives $\times$ 3 seeds (36 runs), and the ArcFace
backbone benchmark uses the same 36-run matrix.  The anti-symmetry ablation
sweeps 9 configurations.  For reference, ArcFace+MiewID training (50 epochs)
takes $\sim$2 hours on MPS.  On a single H100, the end-to-end walltime per
configuration (train + coreset evaluation) is 3.0--7.1 hours (median 3.6 hours)
for Lorentz and 2.2--6.6 hours (median 4.4 hours) for ArcFace, depending on the
backbone.

\clearpage
{\small
\bibliographystyle{unsrt}
\bibliography{references}
}

\end{document}